\documentclass[10pt,twocolumn,letterpaper]{article}

\usepackage{cvpr}
\usepackage{times}
\usepackage{epsfig}
\usepackage{graphicx}
\usepackage{amsmath}
\usepackage{amssymb}
\usepackage{subcaption}
\usepackage[font=small]{caption}
\usepackage{bm}
\usepackage{nomencl}
\usepackage{amsmath}

\DeclareMathOperator*{\argmin}{arg\,min}
\usepackage{amsmath}
\usepackage{subcaption}
\usepackage{amssymb}
\usepackage[export]{adjustbox}

 \makeatletter
\renewcommand{\paragraph}{%
  \@startsection{paragraph}{4}%
  {\z@}{0ex \@plus 1ex \@minus .2ex}{-1em}%
  {\normalfont\normalsize\bfseries}%
}
\makeatother

\usepackage[pagebackref=true,breaklinks=true,letterpaper=true,colorlinks,bookmarks=false]{hyperref}

\cvprfinalcopy 

\def\cvprPaperID{4410} 

\ifcvprfinal\fi
\begin{document}

\title{ Seeing the World in a Bag of Chips}

\author{Jeong Joon Park \hspace{1.5cm} Aleksander Holynski \hspace{1.5cm} Steven M. Seitz \vspace{0.1cm} \\
University of Washington, Seattle\\
{\tt\small \{jjpark7,holynski,seitz\}@cs.washington.edu}
}

\twocolumn[{%
	\renewcommand\twocolumn[1][]{#1}%
	\maketitle
	\begin{center}
		\vspace{-0.7cm}
		\includegraphics[width=0.99\linewidth]{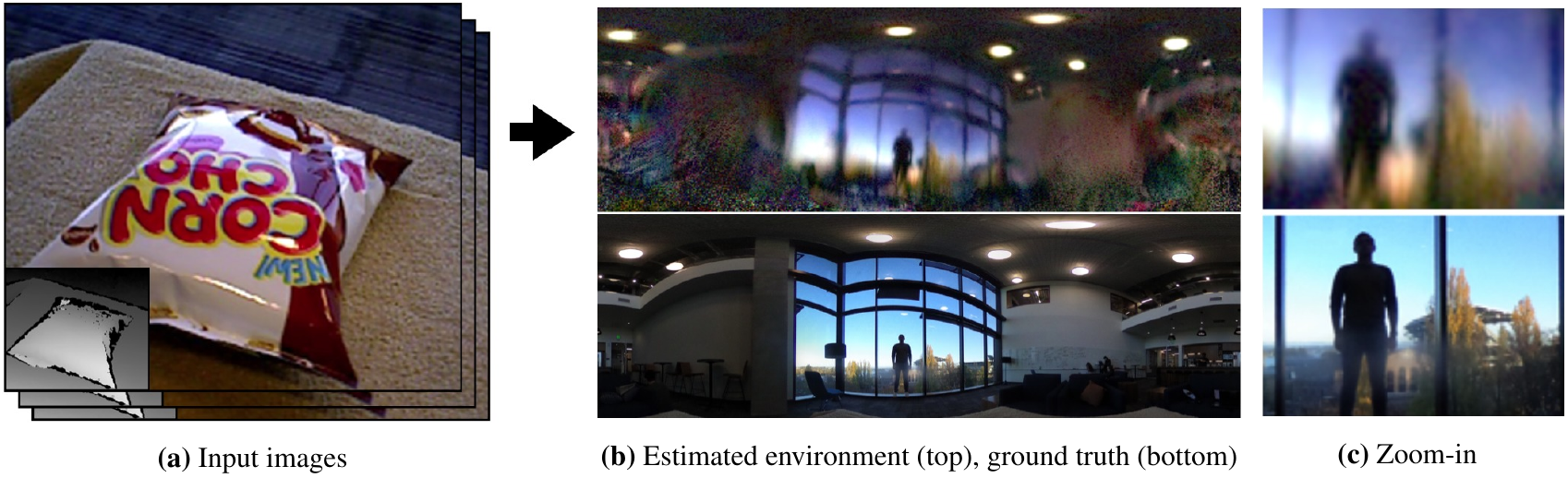}
		\vspace{-0.1cm}
		\captionof{figure}{
From a hand-held RGBD sequence of an object (a), 
we reconstruct an image of the surrounding environment (b, top) that  closely resembles the real environment (b, bottom), entirely from the specular reflections. Note the reconstruction of fine details (c) such as a human figure and trees with fall colors through the window. We use the recovered environment for novel view rendering.$^\dagger$
		}
		\label{fig:teaser}
	\end{center}    
}]

\vspace{-0.3cm}
\begin{abstract}
\vspace{-0.3cm}

We address the dual problems of novel view synthesis and environment reconstruction from hand-held RGBD sensors.
Our contributions include 1) modeling highly specular objects, 2) modeling inter-reflections and Fresnel effects, 
and 3) enabling surface light field reconstruction with the same input needed to reconstruct shape alone.  In cases where scene surface has a strong mirror-like material component, we generate highly detailed environment images, revealing room composition, objects, people, buildings, and trees visible through windows.  Our approach yields state of the art view synthesis techniques, operates on low dynamic range imagery, and is robust to geometric and calibration errors.\let\thefootnote\relax\footnotetext{$^\dagger$\text{Video URL: }\url{https://youtu.be/9t_Rx6n1HGA}}

\end{abstract}
\vspace{-0.5cm}

\section{Introduction}

The glint of light off an object reveals much about its shape and composition -- whether it’s wet or dry, rough or polished, round or flat.  Yet, hidden in the pattern of highlights is also an image of the environment, often so distorted that we don’t even realize it’s there.  Remarkably, images of the shiny bag of chips (Fig.~\ref{fig:teaser}) contain sufficient clues to be able to reconstruct a detailed image of the room, including the layout of lights, windows, and even objects outside that are visible through windows. 

In their {\em visual microphone} work, Davis \etal \cite{Davis14} showed how sound and even conversations can be reconstructed from the minute vibrations visible in a bag of chips.  Inspired by their work, we show that the same bag of chips can be used to reconstruct the environment.  Instead of high speed video, however, we operate on RGBD video, as obtained with commodity depth sensors.

Visualizing the environment is closely connected to the problem of modeling the scene that reflects that environment.
We solve both problems;
beyond visualizing the room, we seek to predict how the objects and scene appear from any new viewpoint — i.e., to virtually explore the scene as if you were there.
This view synthesis problem has a large literature in computer vision and graphics, but several open problems remain.  Chief among them are 1) specular surfaces, 2) inter-reflections, and 3) simple capture.  In this paper we address all three of these problems, 
based on the framework of {\em surface light fields} \cite{wood2000surface}.  

Our environment reconstructions, which we call {\em specular reflectance maps} ({\em SRMs}), represent the distant environment map convolved with the object's specular BRDF.
In cases where the object has strong mirror-like reflections, this SRM provides sharp, detailed features like the one seen in Fig.~\ref{fig:teaser}.  As most scenes are composed of a mixture of materials, each scene has multiple basis SRMs.
We therefore reconstruct a global set of SRMs, together with a weighted material segmentation of scene surfaces. Based on the recovered SRMs, together with additional physically motivated components, we build a neural rendering network capable of faithfully approximating the true surface light field.

A major contribution of our approach is the capability of reconstructing a surface light field with the same 
input needed to compute shape alone \cite{newcombe2011kinectfusion} using an RGBD camera.
Additional contributions of our approach include the ability to operate on regular (low-dynamic range) imagery, 
and applicability to general, non-convex, textured scenes containing multiple objects and both diffuse and specular materials. Lastly, we release RGBD dataset capturing reflective objects to facilitate research on lighting estimation and image-based rendering.

We point out that the ability to reconstruct the reflected scene from images of an object opens up real and valid concerns about privacy.  While our method requires a depth sensor, future research may lead to methods that operate on regular photos.  In addition to educating people on what's possible, our work could facilitate research on privacy-preserving cameras and security techniques that actively identify and scramble reflections.

\section{Related Work}

We review related work in environment lighting estimation and novel-view synthesis approaches for modeling specular surfaces.

\subsection{Environment  Estimation}\label{Sec: environment estimation}

\paragraph{Single-View Estimation}
The most straightforward way to  capture an environment map (image) is via
light probes (e.g., 
a mirrored ball \cite{debevecHDR}) or taking photos with a 360$^\circ$ camera \cite{park2018surface}.  Human eye balls \cite{nishino04} can even serve as light probes when they are present.
For many applications, however, light probes are not available and 
we must rely on existing cues in the scene itself.

Other methods instead study recovering lighting from a photo of a general scene. Because this problem is severely under-constrained, these methods often rely on human inputs \cite{karsch2011rendering,zheng2012interactive} or manually designed ``intrinsic image" priors on illumination, material, and surface properties  \cite{karsch2014automatic,barron2014shape, barron2012shape, bi20151, lombardi2012reflectance}. 

Recent developments in deep learning techniques facilitate data-driven approaches for single view estimation. \cite{gardner2017learning,gardner2019deep,song2019neural,legendre2019deeplight} learn a mapping from a perspective image to a wider-angle 
panoramic image.  
Other methods train models specifically tailored for outdoor scenes \cite{hold2017deep, hold2019deep}. Because the single-view problem is  severely ill-posed, most results are plausible but often non-veridical.
Closely related to our work, Georgoulis \etal \cite{georgoulis2017around} reconstruct higher quality environment images, but under very limiting assumptions; textureless painted surfaces and manual specification of materials and segmentation. 

\paragraph{Multi-View Estimation}
For the special case of planar reflectors,  
layer separation techniques \cite{szeliski2000layer,sinha2012image,xue2015computational,han2017reflection,guo2014robust, jachnik2012real,zhang2018single} enable high quality reconstructions of reflected environments, e.g., from video of a glass picture frame.
Inferring reflections for general, curved surfaces is dramatically harder, even for humans, as the reflected content depends strongly and nonlinearly on surface shape and spatially-varying material properties, 

A number of researchers have sought to recover
{\em low-frequency} lighting from multiple images of curved objects.
\cite{zollhofer2015shading, or2015rgbd, maier2017intrinsic3d} infer spherical harmonics lighting (following \cite{ramamoorthi2001signal})
to refine the surface geometry using principles of shape-from-shading. \cite{richter2016instant} jointly optimizes low frequency  lighting and BRDFs of a reconstructed scene.  While suitable for approximating light source directions, these models don't capture detailed images of the environment.

Wu \etal \cite{wu2015simultaneous}, like us, use a hand-held RGBD sensor to recover lighting and reflectance properties. But the method can only reconstruct a single, floating, convex object, and requires a black background. Dong \etal \cite{dong2014appearance} produces high quality environment images from a video of a single rotating object. This method assumes a laboratory setup with a mechanical rotator, and manual registration of an accurate geometry to their video.  Similarly, Xia \etal \cite{xia2016recovering} use a robotic arm with  calibration patterns to rotate an object. The authors note highly specular surfaces cause trouble, thus limiting their real object samples to mostly rough, glossy materials. In contrast, our  method operates with a hand-held camera for a wide-range of multi-object scenes, and is designed to support specularity.

\subsection{Novel View Synthesis}
Here we focus on  methods capable of modeling {\em specular reflections} from new viewpoints. 

\paragraph{Image-based Rendering}
Light field methods \cite{gortler1996lumigraph,levoy1996light,chen2002light,wood2000surface,davis2012unstructured}
enable highly realistic views of specular surfaces at the expense of 
laborious scene capture from densely sampled viewpoints.
Chen \etal \cite{chen2018deep} regresses surface light fields with neural networks to reduce the number of required views, but requires samples across a full hemisphere captured with a mechanical system. 
Park \etal \cite{park2018surface} avoid dense hemispherical view sampling by applying a parametric BRDF model,
but assume known lighting.

Recent work
applies convolutional neural networks (CNN) to image-based rendering \cite{flynn2016deepstereo,neuralrendering}.
Hedman \etal  \cite{hedman2018deep} replaced the traditional view blending heuristics of IBR systems with a CNN-learned blending weights. Still, novel views are composed of existing, captured pixels, so unobserved specular highlights cannot be synthesized. More recently, \cite{aliev2019neural,thies2019deferred} enhance the traditional rendering pipeline by attaching learned features to 2D texture maps \cite{thies2019deferred} or 3D point clouds \cite{aliev2019neural} and achieve high quality view synthesis results. The features are nonetheless specifically optimized to fit the input views and do not  extrapolate well to novel views. Recent learning-based methods achieve impressive local (versus hemispherical) light field reconstruction from a small set of images \cite{mildenhall2019local,srinivasan2017learning,choi2019extreme,kalantari2016learning,zhou2018stereo}.

\paragraph{BRDF Estimation Methods}

Another way to synthesize novel views is to recover intrinsic surface  reflection functions, known as BRDFs \cite{nicodemus1965directional}. In general, recovering the surface BRDFs is a difficult task, as it involves inverting the complex light transport process. Consequently, existing reflectance capture methods place limits on operating range: e.g., an isolated single object \cite{wu2015simultaneous,dong2014appearance}, known or controlled lighting \cite{park2018surface, debevec1996modeling, lensch2003image, zhou2016sparse,xu2019deep}, single view surface (versus a full 3D mesh) \cite{goldman2010shape,li2018learning}, flash photography \cite{aittala2015two,lee2018practical,nam2018practical}, or spatially constant material \cite{meka2018lime,kim2017lightweight}.

\paragraph{Interreflections}
Very few view synthesis techniques support interreflections. Modeling general multi-object scene requires solving for global illumination (e.g. shadows or interreflections), which is difficult and sensitive to imperfections of real-world inputs \cite{azinovic2019inverse}. Similarly, Lombardi \etal \cite{lombardi2016radiometric} model multi-bounce lighting but with noticeable artifacts and limit their results to mostly uniformly textured objects. Zhang \etal \cite{zhang2016emptying} require manual annotations of light types and locations.

\section{Technical Approach}
Our system takes a video and 3D mesh of a static scene (obtained via Newcombe \etal \cite{newcombe2011kinectfusion}) 
as input and automatically reconstructs an image of the environment along with a scene appearance model that enables novel view synthesis. 
Our approach excels at specular scenes, and accounts for both specular interreflection and Fresnel effects.
A key advantage of our approach is the use of easy, casual data capture from a hand-held camera; we reconstruct the environment map and a surface light field with the same input needed to reconstruct the geometry 
alone, e.g., using \cite{newcombe2011kinectfusion}. 

Section \ref{sec: formul} formulates surface light fields \cite{wood2000surface} and define the specular reflectance map (SRM).
Section \ref{sec: SRM} shows how, 
given geometry and diffuse texture as input, we can jointly recover SRMs and material segmentation through an end-to-end optimization approach. 
Lastly, Section ~\ref{sec:neuralRendering}, describes a {\em scene-specific} neural rendering network that combines recovered SRMs and other rendering components to synthesize realistic novel-view images, with interreflections and Fresnel effects.

\subsection{Surface Light Field Formulation} \label{sec: formul}
We model scene appearance using the concept of a surface light field \cite{wood2000surface}, which defines the color radiance of a surface point in every view direction, given approximate geometry, denoted $\mathcal{G}$ \cite{newcombe2011kinectfusion}.

Formally, the surface light field, denoted $SL$, assigns an RGB radiance value to a ray coming from surface point $\bm{x}$ with outgoing direction $\bm{\omega}$:
    $SL(\bf{x},\bm{\omega})\in \text{RGB}$. 
As is common 
\cite{phong1975illumination,ward1992measuring}, we decompose $SL$ into diffuse (view-independent) and specular (view-dependent) components:
\begin{equation} \label{eq: first}
    SL(\bf{x},\bm{\omega})\approx \mathnormal{D}(\bm{x}) + \mathnormal{S}(\bf{x}, \bm{\omega}).
\end{equation}

\begin{figure} 
\begin{subfigure}[t]{0.48\linewidth}
\includegraphics[width=\linewidth]{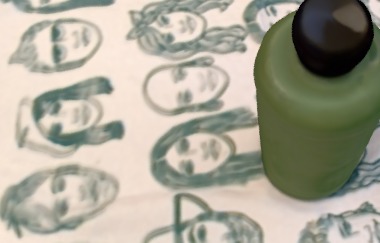}
\caption{Diffuse image $D_P$}
\end{subfigure}
\begin{subfigure}[t]{0.48\linewidth}
\includegraphics[width=\linewidth]{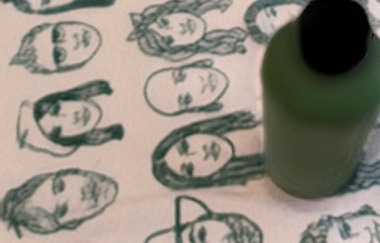}
\caption{Refined Diffuse image $D_P'$}
\end{subfigure}
\caption{The role of diffuse network $u_{\bm{\phi}}$ to correct geometry and texture errors of  RGBD reconstruction. 
The bottle geometry in image (a) is estimated larger than it actually is, and the background textures exhibit ghosting artifacts (faces). The use of the refinement network  corrects these issues (b).
Best viewed digitally.} \label{fig: diffusenet}
\vspace{-0.5cm}
\end{figure}

We compute the diffuse texture $\mathnormal{D}$ for each surface point as the minimum intensity of  across different input views following
\cite{szeliski2000layer,park2018surface}. Because the diffuse component is view-independent, we can then render it from arbitrary viewpoints using the estimated geometry. However, textured 3D reconstructions typically contain errors (e.g., silhouettes are enlarged, as in Fig.~\ref{fig: diffusenet}), so we refine
the rendered texture image using a neural network (Sec. \ref{sec: SRM}).

For the specular component, we define the {specular reflectance map} (SRM)
(also known as {\em lumisphere} \cite{wood2000surface})
and denoted $SR$, 
as a function that maps a reflection ray direction $\bm{\omega}_r$,
defined as the vector reflection of $\bm{\omega}$ about surface normal $\bm{n_x}$
\cite{wood2000surface}
to specular reflectance (i.e., radiance): $SR(\bm{\omega}_r):\Omega \mapsto RGB$,
 where $\Omega$ is a unit hemisphere around the scene center. This model assumes distant environment illumination, although we add support for specular interreflection later in Sec.~\ref{sec:interreflect}.
Note that this model is closely related to prefiltered environment maps \cite{kautz2000unified}, 
used for real-time rendering of specular highlights.

Given a specular reflectance map $SR$, we can render 
the specular image $S$
from a virtual camera as follows:
\begin{equation} \label{eq: spec1}
    S(\bm{x},\bm{\omega}) = V(\bm{x},\bm{\omega}_r;\mathcal{G})\cdot SR(\bm{\omega}_r),
\end{equation} 
where 
$V(\bm{x},\bm{\omega}_r;\mathcal{G})$  is a shadow (visibility) term that is $0$ when the reflected ray $\bm{\omega}_r:=\bm{\omega}-2(\bm{\omega}\cdot \bm{n_x})\bm{n_x}$ from $\bm{x}$ intersects with known geometry $\mathcal{G}$, and $1$ otherwise.

An SRM contains distant environment lighting convolved with a particular specular BRDF. As a result, a single SRM can only accurately describe one surface material. In order to generalize to multiple (and spatially varying) materials, we modify Eq. (\ref{eq: spec1}) by assuming the material at point $\bm{x}$ is a linear combination of $M$ basis materials \cite{goldman2010shape,alldrin2008photometric,zickler2005reflectance}: 
\begin{equation} \label{eq: spec2}
    S(\bm{x},\bm{\omega}) = V(\bm{x},\bm{\omega}_r;\mathcal{G})\cdot \sum_{i=1}^{M} W_i(\bm{x})\cdot 
    SR_{i}(\bm{\omega}_r),
\end{equation}
where $W_i(x) \ge 0$, $\sum_{i=1}^{M}W_i(\bm{x})=1$ and $M$ is user-specified. For each surface point $\bm{x}$, $W_i(\bm{x})$ defines the weight of material basis $i$. We use a neural network to approximate these weights in image-space, as described next.

\subsection{Estimating SRMs and Material Segmentation} \label{sec: SRM}

Given scene shape $\mathcal{G}$ and photos from known viewpoints as input,
we now describe how to 
recover an optimal set of SRMs and material weights. 

Suppose we want to predict a view of the scene from camera $P$ at a pixel $\bm{u}$ that sees surface point $\bm{x_u}$, given known SRMs and material weights. 
We render the diffuse component $D_P(\bm{u})$ from the known diffuse texture $D(\bm{x_u})$, and similarly the blending weight map $W_{P,i}$ from $W_i$ for each SRM using standard rasterization.  A reflection direction image $R_P(\bm{u})$ is obtained by computing per-pixel $\bm{\omega}_r$ values.
We then compute the specular component image $S_P$ by looking up the reflected ray directions $R_P$ in each SRM, and then combining the radiance values using $W_{P,i}$: 
\begin{equation} \label{eq: objective1}
    S_P(\bm{u})=V(\bm{u})\cdot \sum_{i=1}^M W_{P,i}(\bm{u}) \cdot SR_i(R_P(\bm{u})),
\end{equation}
where $V(\bm{u})$ is the visibility term of pixel $\bm{u}$ as used in Eq. (\ref{eq: spec2}).  Each $SR_i$ is stored as a 2D panorama image of resolution 500 x 250 in spherical coordinates.

Now, suppose that SRMs and material weights are unknown;
the optimal SRMs and combination weights minimize the energy $\mathcal{E}$ defined as the sum of differences between the real photos $G$ and the rendered composites of diffuse and specular images $D_P, S_P$ over all input frames $\mathcal{F}$: 
\begin{equation} \label{eq: energy}
\mathcal{E}=\sum_{P\in \mathcal{F}}\mathcal{L}_1 (G_P, D_P + S_P),
\end{equation}
where $\mathcal{L}_1$ is pixel-wise $L1$ loss.

While Eq.~(\ref{eq: energy}) could be minimized directly to obtain  $W_{P,i}$ and $SR_i$, two factors introduce practical difficulties. First, specular highlights tend to be sparse and cover a small percentage of specular scene surfaces. Points on specular surfaces that don't see a highlight are difficult to differentiate from diffuse surface points, thus making the problem of assigning material weights to surface points severely under-constrained. Second, captured geometry is seldom perfect, and misalignments in reconstructed diffuse texture can result in incorrect SRMs. In the remainder of this section, we describe our approach to overcome these limiting factors.

\paragraph{Material weight network.} 
To address the problem of material ambiguity, we pose the material assignment problem as a statistical pattern recognition task. We compute the 2D weight maps $W_{P,i}(\bm{u})$ with a convolutional neural network $w_{\bm{\theta}}$ that learns to map a diffuse texture image patch to the blending weight of $i$th material:
    $W_{P,i}=w_{\bm{\theta}}(D_P)_i.$
This network learns correlations between diffuse texture and material properties (i.e., shininess), and is trained on each scene by jointly optimizing the network weights and SRMs to reproduce the input images. 

Since $w_{\bm{\theta}}$ predicts material weights in image-space, and therefore per view, we introduce a view-consistency regularization function $\mathcal{V}(W_{P_1}, W_{P_2})$ penalizing the pixel-wise $L1$ difference in the predicted materials between a pair of views when cross-projected to each other (i.e., one image is warped to the other using the known geometry and pose).

\paragraph{Diffuse refinement network.} Small errors in geometry and calibration, as are typical in scanned models, cause misalignment and ghosting artifacts in the texture reconstruction $D_P$.  Therefore, we introduce
a refinement network $u_{\bm{\phi}}$ to correct these errors (Fig. \ref{fig: diffusenet}). We replace $D_P$ with the refined texture image: $D_P'=u_{\bm{\phi}}(D_P)$. Similar to the material weights, we penalize the inconsistency of the refined diffuse images across viewpoints using $\mathcal{V}(D_{P_1}',D_{P_2}').$ Both networks $w_{\bm{\theta}}$ and $u_{\bm{\phi}}$ follow the encoder-decoder architecture with residual connections \cite{johnson2016perceptual,he2016deep}, while $w_{\bm{\theta}}$ has lower number of parameters.
We refer readers to supplementary for more details.

\paragraph{Robust Loss.} Because a pixel-wise loss alone is not robust to misalignments, 
we define the image distance metric $\mathcal{L}$ as a combination of pixel-wise $L1$ loss, perceptual loss $\mathcal{L}_{p}$ computed from feature activations of a pretrained network \cite{chen2017photographic}, and adversarial loss \cite{goodfellow2014generative,isola2017image}. 
Our total loss, for a pair of images $I_1, I_2$, is:
\begin{equation}    
\begin{aligned}
\mathcal{L}(I_1, I_2;d) &= \lambda_{1}\mathcal{L}_1(I_1, I_2)+\lambda_{p}\mathcal{L}_{p}(I_1, I_2)\\&+\lambda_{G}\mathcal{L}_{G}(I_1, I_2;d),
\end{aligned}
\end{equation}
where $d$ is the discriminator, and $\lambda_{1} = 0.01$, $\lambda_{p} = 1.0$, and $\lambda_{G}=0.05$ are balancing coefficients.
The neural network-based perceptual and adversarial loss are effective because they are robust to image-space misalignments caused by errors in the estimated geometry and poses.

Finally, we add a sparsity term on the specular image $||S_P||_1$ to regularize the specular component from containing colors from the diffuse texture.
 
\begin{figure}
\includegraphics[width=\linewidth]{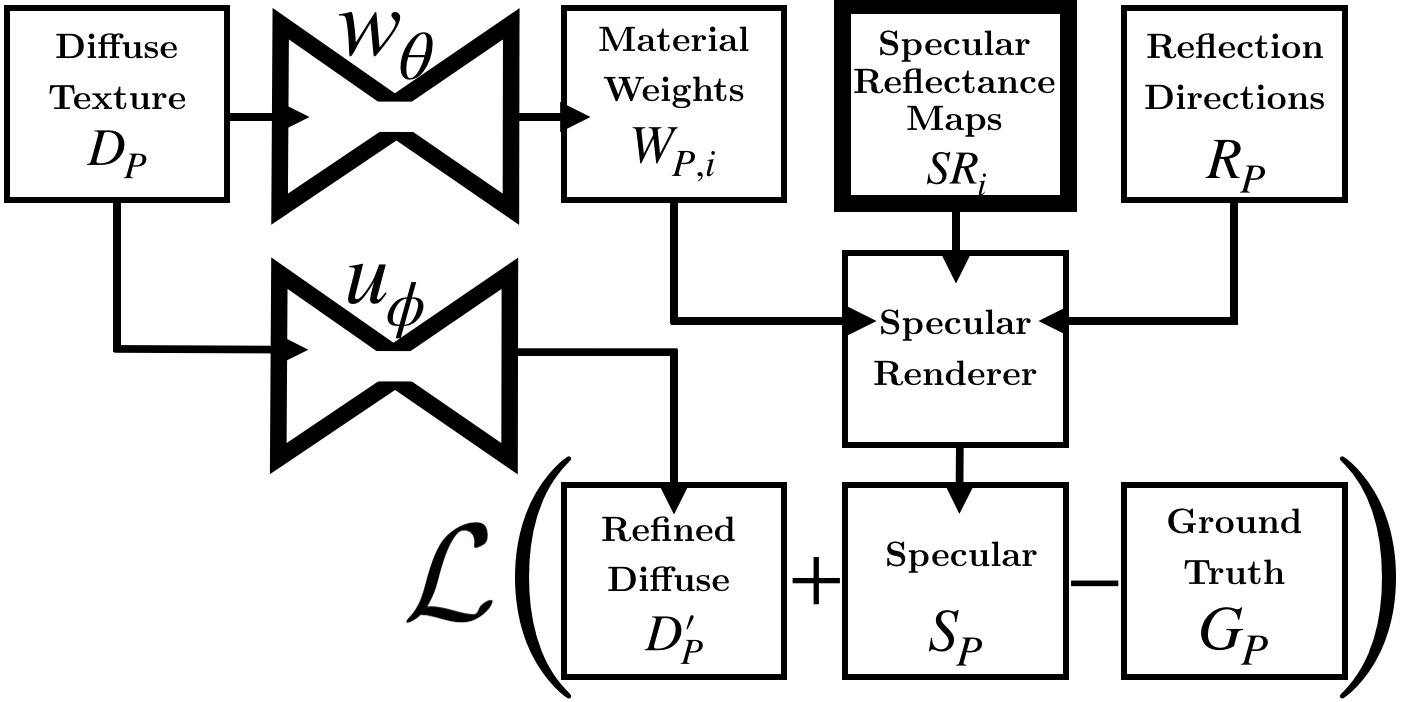}
\caption{The components of our SRM estimation pipeline (optimized parameters shown in bold). We predict a view by adding refined diffuse texture $D'_P$ (Fig.~\ref{fig: diffusenet}) and the specular image $S_P$. $S_P$ is computed, for each pixel, by looking up the basis SRMs ($SR_i$'s) with surface reflection direction $R_P$ and blending them with  weights $W_{P,i}$ obtained via network $w_\theta$. The loss between the predicted view and ground truth  $G_P$ is backpropagated to jointly optimize the SRM pixels and network weights.
 }\label{fig: overview}
\end{figure}

Combining all elements, we get the final loss function:
\begin{equation} \label{eq: objective2}
\begin{aligned}
    &SR^*,\bm{\theta}^*,\bm{\phi}^* =\argmin_{SR, \bm{\theta}, \bm{\phi}} \max_d\sum_{P\in \mathcal{F}} \mathcal{L} (G_P, D_P'+S_P;d) \\& + \lambda_S||S_P||_1+\lambda_V\mathcal{V}(W_P,W_{P_r})+\lambda_T\mathcal{V}(D_P',D_{P_r}'),
\end{aligned}
\end{equation}
where $P_r$ is a randomly chosen frame in the same batch
 with $P$ during each stochastic gradient descent step. $\lambda_S$, $\lambda_T$ and $\lambda_V$ are set to 1e-4. An overview diagram is shown in Fig.~\ref{fig: overview}. Fig. \ref{fig: lighting} shows that the optimization discovers coherent material regions and a detailed environment image.

\begin{figure}
\begin{subfigure}[t]{0.325\linewidth}
\includegraphics[width=0.9\linewidth]{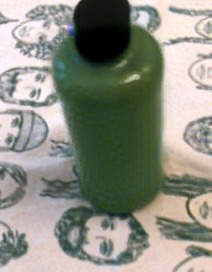}
\caption{W/O Interreflections}
\end{subfigure}
\begin{subfigure}[t]{0.325\linewidth}
\includegraphics[width=0.9\linewidth]{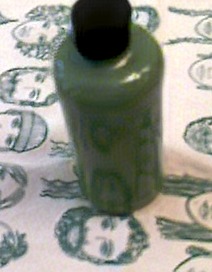}
\caption{With Interreflections}
\end{subfigure}
\begin{subfigure}[t]{0.325\linewidth}
\includegraphics[width=0.9\linewidth]{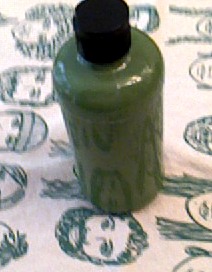}
\caption{Ground Truth}
\end{subfigure}

\vspace{0.1cm}
\begin{subfigure}[t]{0.325\linewidth}
\includegraphics[width=0.95\linewidth]{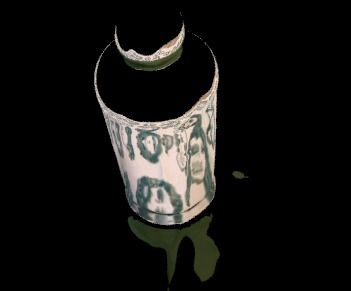}
\caption{FBI}
\end{subfigure}
\begin{subfigure}[t]{0.325\linewidth}
\includegraphics[width=0.95\linewidth]{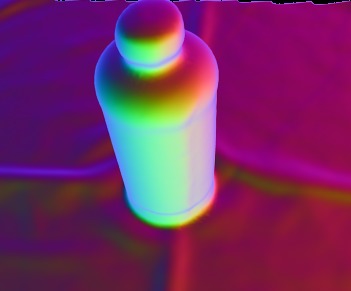}
\caption{$R_P$}
\end{subfigure}
\begin{subfigure}[t]{0.325\linewidth}
\includegraphics[width=0.95\linewidth]{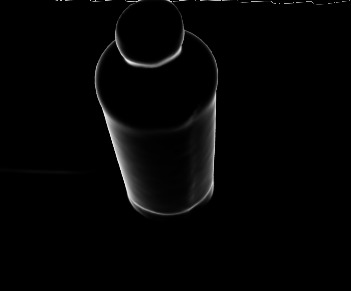}
\caption{Fresnel}
\end{subfigure}

\caption{Modeling interreflections. First row shows images of an unseen viewpoint rendered by a network trained with direct (a) and with interreflection + Fresnel models (b), compared to ground truth (c).  Note accurate interreflections on the bottom of the green bottle (b). 
(d), (e), and (f) show first-bounce image (FBI), reflection direction image ($R_P$), and Fresnel coefficient image (FCI), respectively. Best viewed digitally.
\label{fig: interreflections}}
\vspace{-0.5cm}
\end{figure}

\begin{figure*}
\includegraphics[trim=0 20 0 10,clip,width=0.135\linewidth]{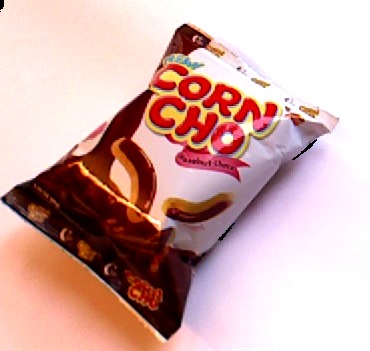}
\includegraphics[trim=0 20 0 1
10,clip,width=0.135\linewidth]{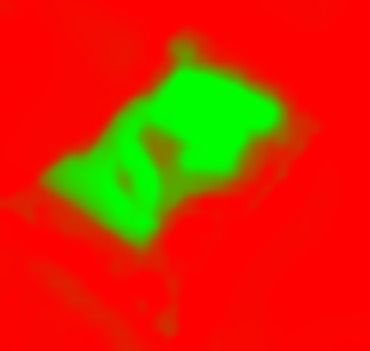}
\includegraphics[width=0.357\linewidth]{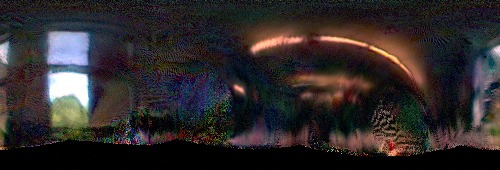}
\includegraphics[width=0.357\linewidth]{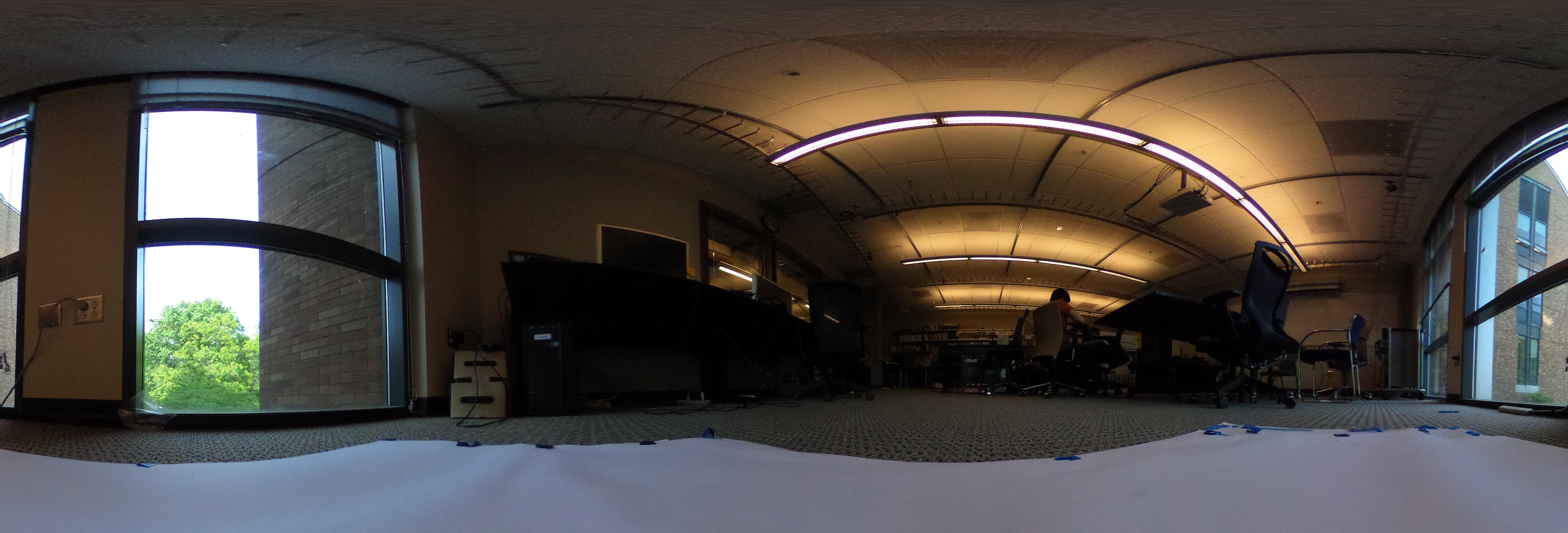}

\begin{subfigure}[t]{0.135\linewidth}
\includegraphics[trim=0 5 0 0,clip,width=\linewidth]{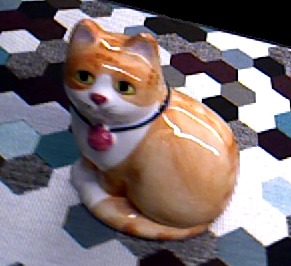}
\caption{Input Video}
\end{subfigure}
\begin{subfigure}[t]{0.135\linewidth}
\includegraphics[trim=0 5 0 0,clip,width=\linewidth]{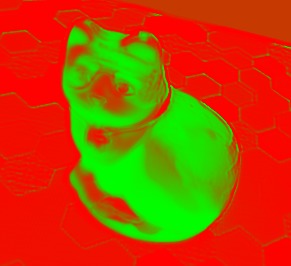}
\caption{Material Weights}
\end{subfigure}
\begin{subfigure}[t]{0.357\linewidth}
\includegraphics[width=\linewidth]{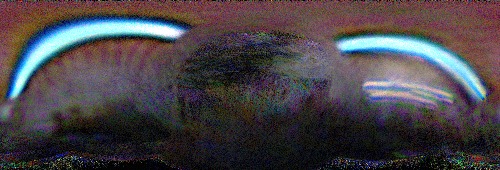}
\caption{Recovered SRM}
\end{subfigure}
\begin{subfigure}[t]{0.357\linewidth}
\includegraphics[width=\linewidth]{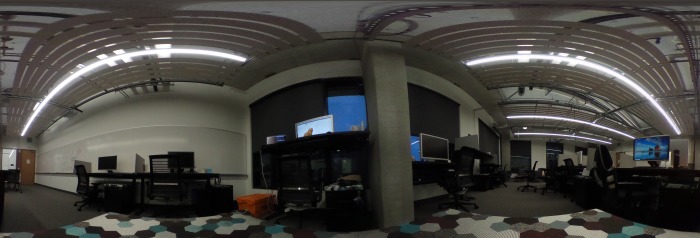}
\caption{Ground Truth}
\end{subfigure}

\begin{subfigure}[t]{0.3355\linewidth}
\includegraphics[width=\linewidth]{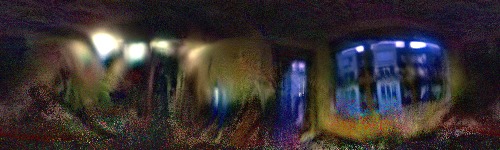}
\caption{Recovered SRM}
\end{subfigure}
\begin{subfigure}[t]{0.3355\linewidth}
\includegraphics[width=\linewidth]{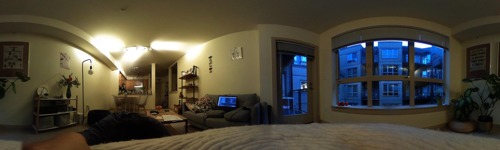}
\caption{Ground Truth}
\end{subfigure}
\begin{subfigure}[t]{0.1585\linewidth}
\includegraphics[width=\linewidth]{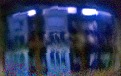}
\caption{Zoom-in(ours)}
\end{subfigure}
\begin{subfigure}[t]{0.1585\linewidth}
\includegraphics[width=\linewidth]{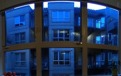}
\caption{Zoom-in(GT)}
\end{subfigure}

\caption{Sample results of recovered SRMs and material weights. Given input video frames (a), we recover global SRMs (c) and their linear combination weights (b) from the optimization of Eq. (\ref{eq: objective2}). The scenes presented here have two material bases, visualized with red and green channels. Estimated SRMs (c) corresponding to the shiny object surface (green channel) correctly capture the light sources of the scenes, shown in the reference panorama images (d). For both scenes the SRMs corresponding to the red channel is mostly black, thus not shown, as the surface is mostly diffuse. The recovered SRM of (c) overemphasizes blue channel due to oversaturation in input images. Third row shows estimation result from a video of the same bag of chips (first row) under different lighting. Close inspection of the recovered environment (g) reveals many scene details, including floors in a nearby building visible through the window.
} \label{fig: lighting}
\end{figure*}

\begin{figure*}
\hfill
\begin{subfigure}[t]{0.095\linewidth}
\includegraphics[width=\linewidth]{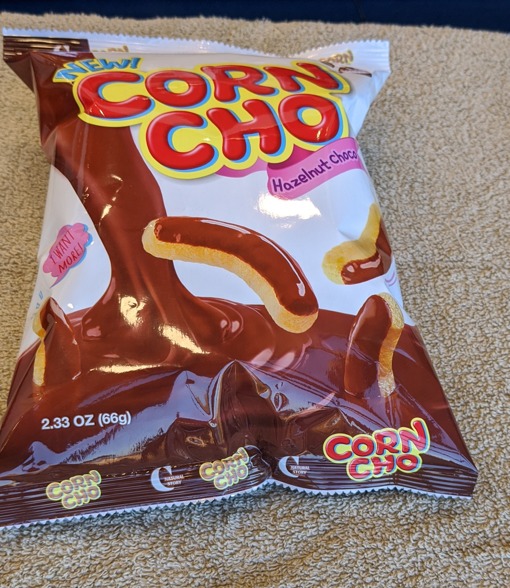}
\caption{Input}
\end{subfigure}
\hfill
\begin{subfigure}[t]{0.22\linewidth}
\includegraphics[width=\linewidth]{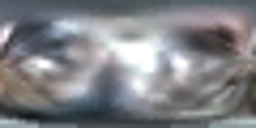}
\caption{Legendre \etal \cite{legendre2019deeplight} }
\end{subfigure}
\hfill
\begin{subfigure}[t]{0.22\linewidth}
\includegraphics[width=\linewidth]{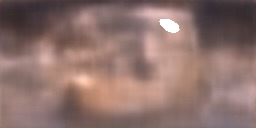}
\caption{Gardner \etal \cite{gardner2017learning} }
\end{subfigure}
\hfill
\begin{subfigure}[t]{0.22\linewidth}
\includegraphics[width=\linewidth]{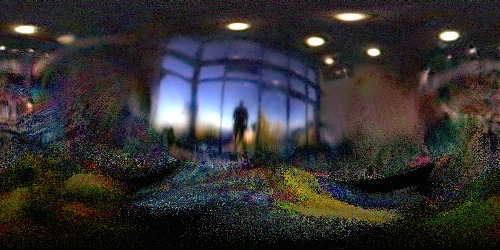}
\caption{Our Result }
\end{subfigure}
\hfill
\begin{subfigure}[t]{0.22\linewidth}
\includegraphics[width=\linewidth]{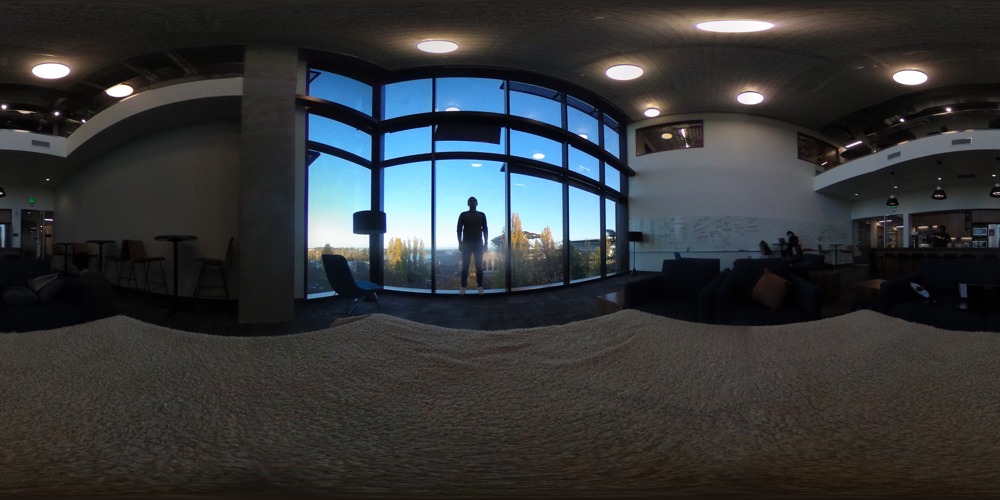}
\caption{Ground Truth}
\end{subfigure}
\hfill

\begin{subfigure}[t]{0.13\linewidth}
\includegraphics[width=0.85\linewidth]{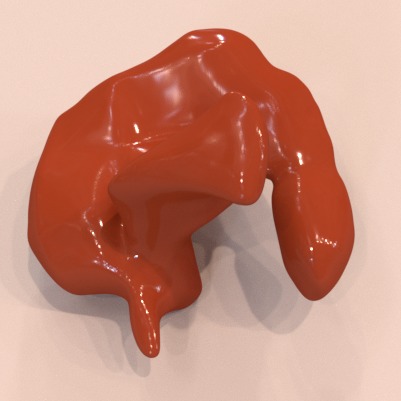}
\caption{Synthetic Scene}
\end{subfigure}
\begin{subfigure}[t]{0.27\linewidth}
\includegraphics[width=\linewidth]{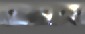}
\caption{Lombardi \etal \cite{lombardi2016radiometric} }
\end{subfigure}
\hfill
\begin{subfigure}[t]{0.27\linewidth}
\includegraphics[width=\linewidth]{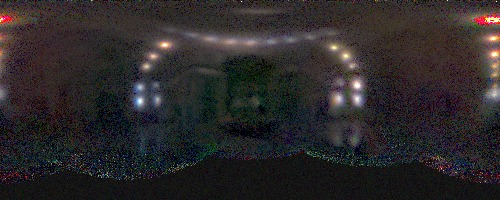}
\caption{Our Result }
\end{subfigure}
\hfill
\begin{subfigure}[t]{0.27\linewidth}
\includegraphics[width=\linewidth]{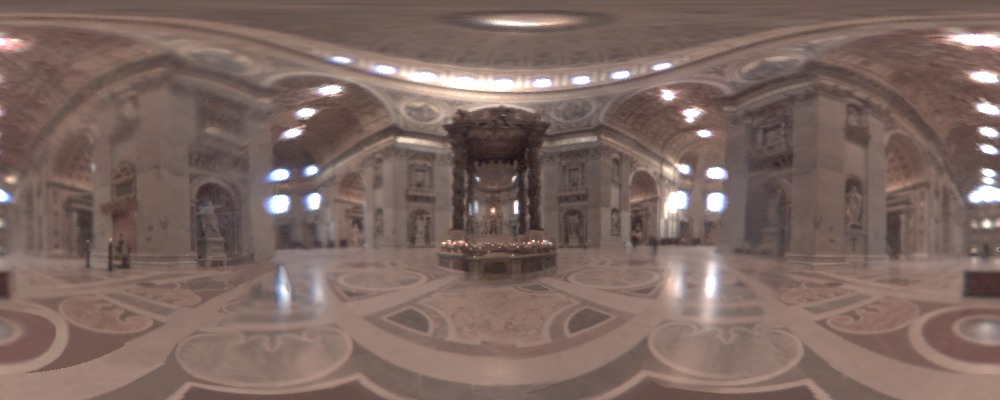}
\caption{Ground Truth }
\end{subfigure}
\hfill

\caption{Comparisons with existing single-view and multi-view based environment estimation methods. Given a single image (a),
Deeplight \cite{legendre2019deeplight}
(b), and Gardner \etal \cite{gardner2019deep} (c), do not produce accurate environment reconstructions, relative to what we obtain from an RGBD video (d) which better matches ground truth (e).
Additionally, from a video sequence and noisy geometry of a synthetic scene (f), our method (h) more accurately recovers the surrounding environment (i) compared to Lombardi \etal (g). 
} \label{fig: compairson lighting}
\vspace{-0.5cm}
\end{figure*}

\subsection{Novel-View Neural Rendering}
\label{sec:neuralRendering} 

With reconstructed SRMs and material weights, we can synthesize specular appearance from any desired viewpoint via Eq. (\ref{eq: spec1}).
However, 
while the approach detailed in Sec.~\ref{sec: SRM} reconstructs high quality SRMs, the renderings 
often lack realism 
(shown in supplementary), due to two factors.
First, errors in geometry and camera pose can sometimes lead to weaker reconstructed highlights.
Second, the SRMs do not model more complex light transport effects such as interreflections or Fresnel reflection. 
This section describes how we train a network to 
address these two limitations, yielding more realistic results.

Simulations only go so far, and computer renderings will never be perfect.  
In principle, you could train a CNN to render images as a function of viewpoint directly, 
training on actual photos.
Indeed, several recent neural rendering methods adapt image translation \cite{isola2017image}
to learn mappings from projected point clouds \cite{neuralrendering,pittaluga2019revealing,aliev2019neural} or a UV map image \cite{thies2019deferred} to a photo.
However, these methods struggle to extrapolate far away from the input views because their networks don't have built-in physical models of specular light transport.

Rather than treat the rendering problem as a black box, we arm the neural renderer with knowledge
of physics -- in particular, diffuse, specular, interreflection, and Fresnel reflection, to use in learning
how to render images.
Formally, we introduce an adversarial neural network-based generator $g$ and discriminator $d$ to render realistic photos.  $g$ takes as input our best prediction of diffuse $D_P$ and specular $S_P$ components for the current view
(obtained from Eq. (\ref{eq: objective2})), along with interreflection and Fresnel terms $FBI$, $R_P$, and $FCI$ that will be defined later in this section.

Consequently, the generator $g$ receives $C_P=(D_P,S_P,$ $FBI, R_P, FCI)$ as input and outputs a prediction of the view, while the discriminator $d$ scores its realism.  We use the combination of pixelwise $L_1$, perceptual loss $L_{p}$ \cite{chen2017photographic}, and the adversarial loss \cite{isola2017image} as described in Sec. \ref{sec: SRM}:
\vspace{-0.4cm}

\begin{equation} \label{eq: GAN}
g^* = \argmin_g \max_d \lambda_G\bar{\mathcal{L}}_{G}(g,d) + \lambda_p \bar{\mathcal{L}}_{p}(g) + \lambda_{1} \bar{\mathcal{L}}_{1}(g),
\vspace{-0.2cm}
\end{equation}
where $\bar{\mathcal{L}}_{p}(g)=\frac{1}{|\mathcal{F}|}\sum_{P\in\mathcal{F}} \mathcal{L}_{p}(g(C_P),G_P)$ is the mean of perceptual loss across all input images, and $\mathcal{L}_{G}(g,d)$ and $\bar{\mathcal{L}}_{1}(g)$ are similarly defined as an average loss across frames.  Note that this renderer $g$ is {\em scene specific}, trained
only on images of a particular scene to extrapolate new views of that same scene, as commonly done in the neural rendering community \cite{neuralrendering,thies2019deferred,aliev2019neural}.

\paragraph{Modeling Interreflections and Fresnel Effects} \label{sec:interreflect}

Eq.~(\ref{eq: spec1}) models only the direct illumination of each surface point by the environment, neglecting interreflections. 
While modeling full, global, diffuse + specular light transport is intractable, we can approximate first order interreflections by ray-tracing a first-bounce image (FBI) as follows.  
For each pixel $\bf u$ in the virtual viewpoint to be rendered, cast a ray from the camera center through $\bf u$.  If we pretend for now that every scene surface is a perfect mirror, that ray will bounce potentially multiple times and intersect multiple surfaces.  Let ${\bf x}_2$ be the second point of intersection of that ray with the scene.  Render the
pixel at $\bf u$ in FBI with the diffuse color of ${\bf x}_2$, or with black if there is no second intersection (Fig.~\ref{fig: interreflections}(d)). 

Glossy (imperfect mirror) interreflections can be modeled by 
convolving the FBI with the BRDF.  Strictly speaking, however, the
interreflected image should be filtered in the {\em angular domain} \cite{ramamoorthi2001signal}, rather than image space,
i.e., convolution of incoming light following the specular lobe whose center is the reflection ray direction $\bm{\omega}_r$.
Given $\bm{\omega}_r$, angular domain convolution can be approximated in image space by convolving the FBI image weighted by $\bm{\omega}_r$.
However, because we do not know the specular kernel, we let the network infer the weights using $\bm{\omega}_r$ as a guide.
We encode the $\bm{\omega}_r$ for each pixel as a three-channel image $R_P$ (Fig.~\ref{fig: interreflections}(e)).

Fresnel effects make highlights stronger at near-glancing view angles 
and are important for realistic rendering.
Fresnel coefficients are approximated following \cite{schlick1994inexpensive}:
$R(\alpha)=R_0+(1-R_0)(1-cos\alpha)^5,$ 
where $\alpha$ is the angle between the surface normal and the camera ray, and $R_0$ is a material-specific constant. We compute a Fresnel coefficient image (FCI), where each pixel contains $(1-cos\alpha)^5$, and provide it to the network as an additional input, shown in Fig.~\ref{fig: interreflections}(f).

In total, the rendering components $C_P$ are now composed of five images: diffuse and specular images, FBI image, $R_P$, and FCI. $C_P$ is then given as input to the neural network, and our network weights are optimized as in Eq. (\ref{eq: GAN}). Fig. \ref{fig: interreflections} shows the effectiveness of the additional three rendering components for modeling interreflections.
\vspace{-0.05cm}

\subsection{Implementation Details}
We follow \cite{johnson2016perceptual} for the generator network architecture, use the PatchGAN discriminator \cite{isola2017image}, and employ the loss of LSGAN \cite{mao2017least}. We use ADAM \cite{kingma2014adam} with learning rate 2e-4 to optimize the objectives. Data augmentation was essential for viewpoint generalization, by applying random rotation, translation, flipping, and scaling to each input and output pair. More details can be found in supplementary.

\begin{figure*}

\begin{subfigure}[t]{\linewidth}

\hfill
\includegraphics[width=0.193\linewidth]{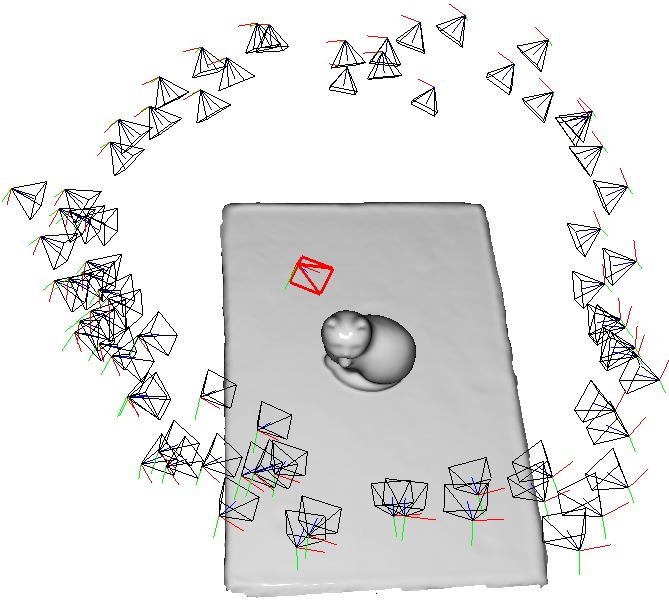}
\hfill
\includegraphics[width=0.193\linewidth]{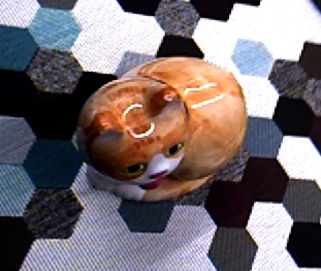}
\hfill
\includegraphics[width=0.193\linewidth]{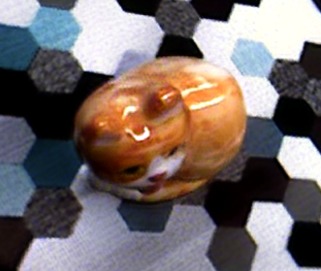}
\hfill
\includegraphics[width=0.193\linewidth]{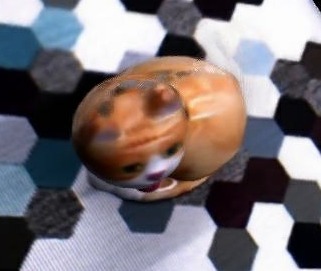}
\hfill
\includegraphics[width=0.193\linewidth]{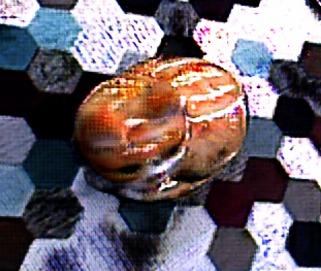}
\hfill
\end{subfigure}

\vspace{0.1cm}

\hfill
\begin{subfigure}[t]{0.193\linewidth}
\includegraphics[width=\linewidth]{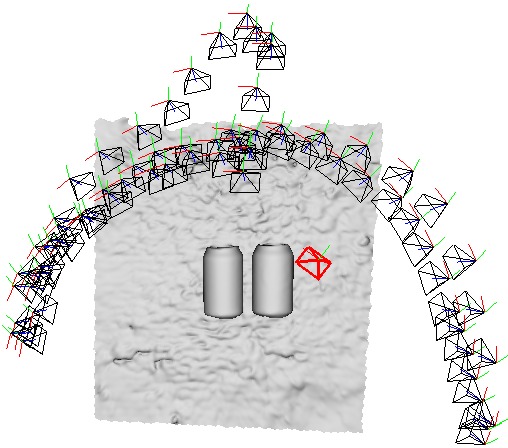}
\caption{Camera Trajectory}
\end{subfigure}
\hfill
\begin{subfigure}[t]{0.193\linewidth}
\includegraphics[width=\linewidth]{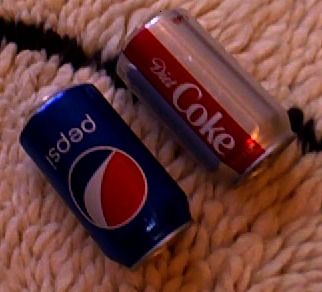}
\caption{Reference Photo}
\end{subfigure}
\hfill
\begin{subfigure}[t]{0.193\linewidth}
\includegraphics[width=\linewidth]{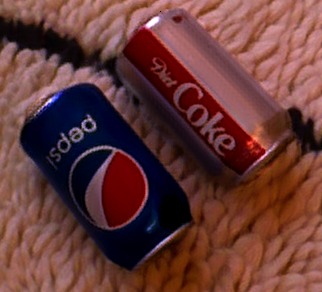}
\caption{Ours}
\end{subfigure}
\hfill
\begin{subfigure}[t]{0.193\linewidth}
\includegraphics[width=\linewidth]{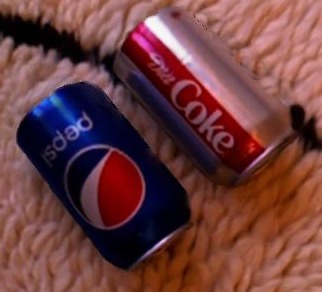}
\caption{DeepBlending \cite{hedman2018deep}}
\end{subfigure}
\hfill
\begin{subfigure}[t]{0.193\linewidth}
\includegraphics[width=\linewidth]{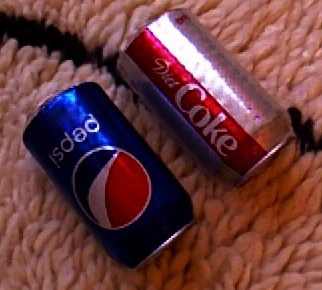}
\caption{Thies \etal \cite{thies2019deferred}}
\end{subfigure}
\hfill

\caption{View extrapolation to extreme viewpoints. We evaluate novel view synthesis on test views (red frusta) that are furthest from the input views (black frusta) (a). The view predictions of DeepBlending \cite{hedman2018deep} and Thies \etal \cite{thies2019deferred} (d,e) are notably different from the reference photographs (b), e.g., missing highlights on the back of the cat, and incorrect highlights at the bottom of the cans. Thies \etal \cite{thies2019deferred} shows severe artifacts, likely because their learned UV texture features overfits to the input views, and thus cannot generalize to very different viewpoints. Our method (c) produces images with highlights appearing at correct locations.}\label{fig: extrapolate}
\end{figure*}

\subsection{Dataset}
We captured ten sequences of RGBD video with a hand-held Primesense  depth camera, featuring a wide range of materials, lighting, objects, environments, and camera paths. The length of each sequence ranges from 1500 to 3000 frames, which are split into train and test frames. Some of the sequences were captured such that the test views are very far from the training views, making them ideal for benchmarking the extrapolation abilities of novel-view synthesis methods. Moreover, many of the sequences come with ground truth HDR environment maps to facilitate future research on environment estimation. Further capture and data-processing details are in supplementary.

\section{Experiments} \label{sec:exp}
 We describe experiments to test our system's ability to estimate images of the environment and synthesize novel viewpoints, and
ablation studies to characterize the 
factors that most contribute to system performance.

We compare our approach to several state-of-the-art methods:
recent single view lighting estimation methods (DeepLight \cite{legendre2019deeplight}, Gardner \etal \cite{gardner2017learning}), an RGBD video-based lighting and material reconstruction method  \cite{lombardi2016radiometric}, an IR-based BRDF estimation method  \cite{park2018surface} (shown in supplementary), and two leading view synthesis methods capable of handling specular highlights -- DeepBlending \cite{hedman2018deep} and Deferred Neural Rendering (DNS) \cite{thies2019deferred}. 

\subsection{Environment Estimation}
Our computed SRMs
demonstrate our system's ability to infer detailed images of the environment 
from the pattern and motion of specular highlights on an object.
For example from \ref{fig: lighting}(b), we can see the general layout of the living room, and even count the number of floors in buildings visible through the window. Note that the person capturing the video does not appear in the environment map because he is constantly moving. 
The shadow of the moving person, however, causes artifacts, e.g. the fluorescent lighting in the first row of Fig.~\ref{fig: lighting} is not fully reconstructed.

Compared to state-of-the-art single view estimation methods \cite{legendre2019deeplight,gardner2017learning}, our method produces a more accurate 
image of the environment, as shown in Fig.~\ref{fig: compairson lighting}. Note our reconstruction shows a person standing near the window and autumn colors in a tree visible through the window.

We compare with a multi-view RGBD based method \cite{lombardi2016radiometric} on a synthetic scene containing a red object, obtained from the authors. As in \cite{lombardi2016radiometric}, we estimate lighting from the known geometry with added noise and a video of the scene rendering, 
but produce more accurate results (Fig.~\ref{fig: compairson lighting}).

\subsection{Novel View Synthesis}\label{sec: NVS}
We recover specular reflectance maps and train a generative network for each video sequence. The trained model is then used to generate novel views from held-out views. 

In the supplementary, we show novel view generation results for different scenes, along with the intermediate rendering components and ground truth  images.  As view synthesis results are better shown in video form, we strongly encourage readers to watch the {\em supplementary video}. 

\begin{figure}
\includegraphics[width=0.99\linewidth]{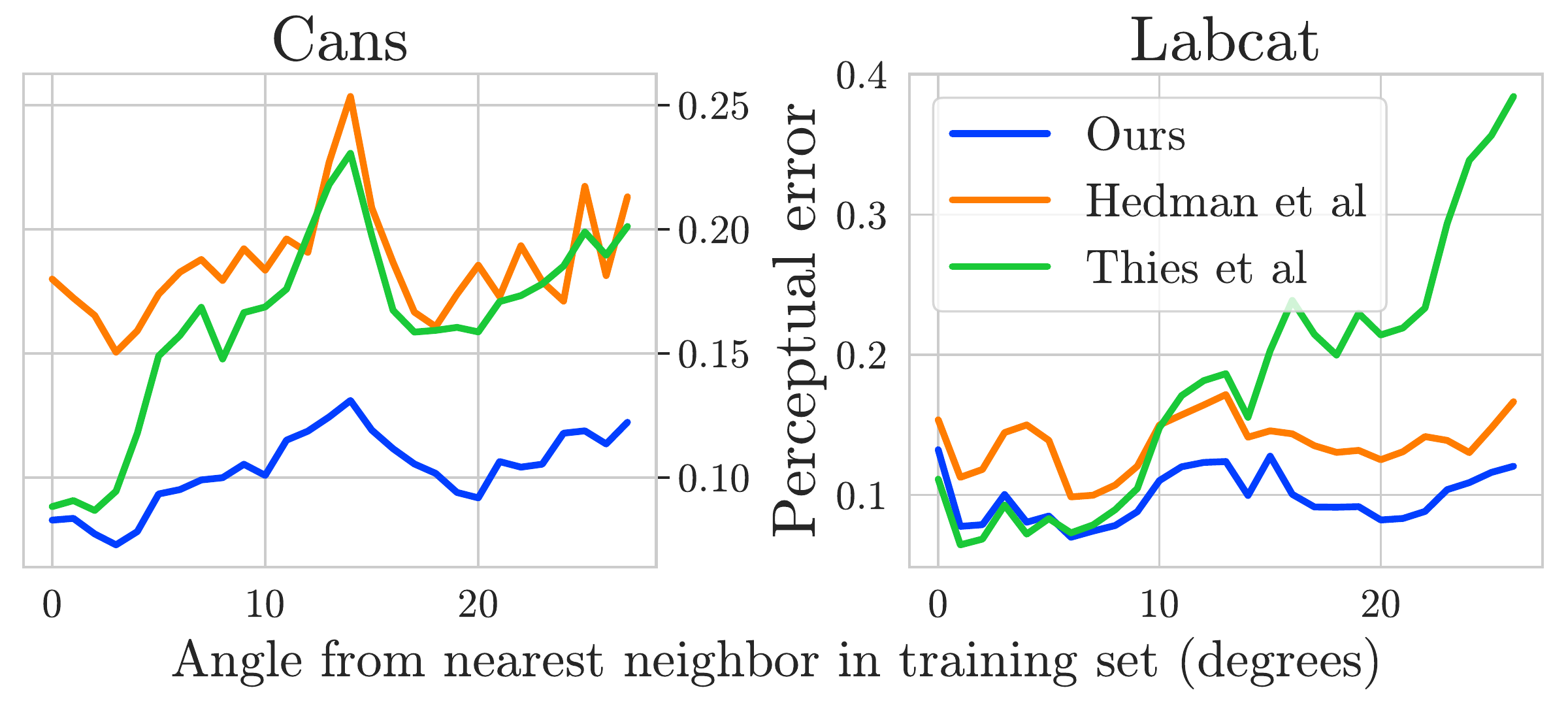}
\caption{Quantitative comparisons for novel view synthesis. We plot the perceptual loss \cite{zhang2018unreasonable} between a novel view rendering and the  ground truth test image as a function of its distance to the nearest training view (measured in angle between the view vectors). We compare our method with two leading NVS methods \cite{hedman2018deep,thies2019deferred} on two scenes. On average, our results have lowest error. 
} \label{fig: graph}
\end{figure}
\begin{figure}
\begin{subfigure}[t]{\linewidth}

\hfill
\begin{subfigure}[t]{0.49\linewidth}
\includegraphics[width=\linewidth]{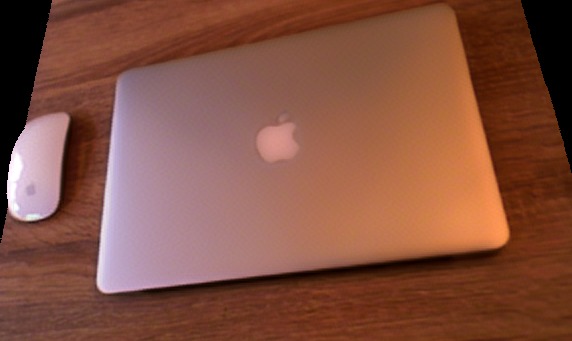}
\caption{Synthesized Novel-view}
\end{subfigure}
\hfill
\begin{subfigure}[t]{0.49\linewidth}
\includegraphics[width=\linewidth]{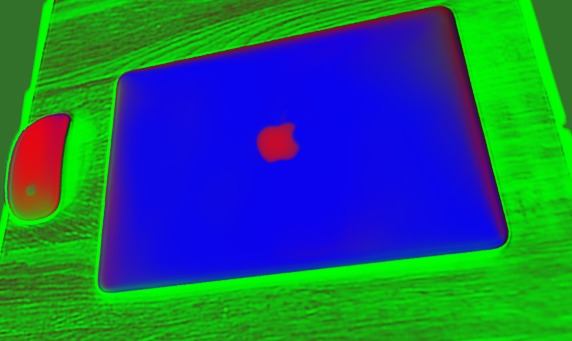}
\caption{Material Weights}
\end{subfigure}
\hfill

\end{subfigure}

\caption{Image (a) shows a synthesized novel view using neural rendering (Sec. \ref{sec:neuralRendering}) of a scene with multiple glossy materials. The spatially varying materials (SRM blending weights) of the wooden tabletop and the laptop are accurately estimated by our algorithm (Sec. \ref{sec: SRM}), as visualized in (b).
}\label{fig: glossy}
\end{figure}

\begin{figure}
\begin{subfigure}[t]{\linewidth}

\hfill
\begin{subfigure}[t]{0.49\linewidth}
\includegraphics[width=\linewidth]{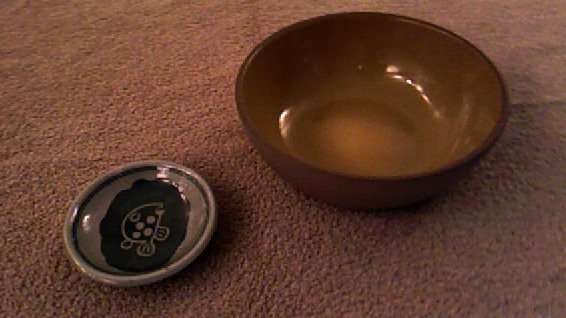}
\caption{Ground Truth}
\end{subfigure}
\hfill
\begin{subfigure}[t]{0.49\linewidth}
\includegraphics[width=\linewidth]{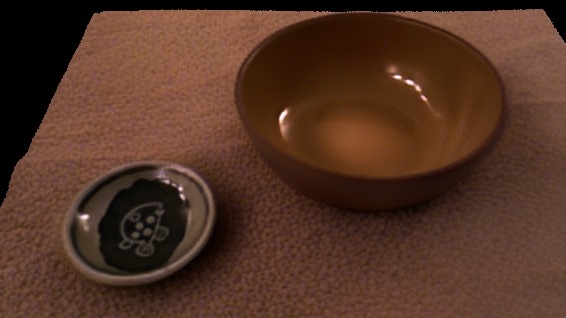}
\caption{Synthesized Novel-view}
\end{subfigure}
\hfill

\end{subfigure}

\caption{Concave surface reconstruction. The appearance of highly concave bowls is realistically reconstructed by our system. The rendered result (b) captures both occlusions and highlights of the ground truth (a).}\label{fig: concave}
\end{figure}

\paragraph{{Novel View Extrapolation}}
Extrapolating novel views far from the input range is 
particularly challenging for scenes with reflections. To test the operating range of our and other recent view synthesis results, we study how the quality of view prediction degrades as a function of the distance to the nearest input images (in difference of viewing angles) (Fig.~\ref{fig: graph}). We measure prediction quality with perceptual loss \cite{zhang2018unreasonable}, which is known to be more robust to shifts or misalignments, against the ground truth test image taken from same pose. We use two video sequences both containing highly reflective surfaces and with large differences in train and test viewpoints. We focus our attention on parts of the scene which exhibit significant view-dependent effects. That is, we mask out the diffuse backgrounds and measure the loss on only central objects of the scene. We compare our method with DeepBlending \cite{hedman2018deep} and Thies \etal \cite{thies2019deferred}. The quantitative (Fig.~\ref{fig: graph}) and qualitative (Fig.~\ref{fig: extrapolate}) results show that our method is able to produce more accurate images of the scene from extrapolated viewpoints.

\subsection{Robustness}
Our method is robust to 
various scene configurations, such as scenes containing multiple objects (Fig. \ref{fig: extrapolate}), spatially varying materials (Fig.~\ref{fig: glossy}), and concave surfaces (Fig.~\ref{fig: concave}). In the supplementary, we study how the loss functions and surface roughness affect our results.

\section{Limitations and Future work}
Our approach relies on the reconstructed mesh obtained from fusing depth images of consumer-level depth cameras and thus fails for surfaces out of the operating range of these cameras, e.g., thin, transparent, or mirror surfaces. Our recovered environment images are filtered by the surface BRDF; separating these two factors is an interesting topic of future work, perhaps via data-driven deconvolution (e.g. \cite{xu2014deep}).
Last, reconstructing a room-scale photorealistic appearance model remains a major open challenge.

\section*{Acknowledgement}
\noindent This work was supported by funding from Facebook, Google, Futurewei, and the
UW Reality Lab.

{\small
\bibliographystyle{ieee_fullname}
\bibliography{sample-bibliography}
}

\appendix

\section*{Supplementary}
\section{Overview}
In this document we provide additional experimental results and extended technical details to supplement the main submission. We first discuss the effects on the output of the system made by changes in the loss functions (Sec.~\ref{sec: loss}), scene surface characteristics (surface roughness) (Sec.~\ref{sec: roughness}), and number of material bases (Sec.~\ref{sec: basis}). We then showcase our system's ability to model the Fresnel effect (Sec.~\ref{sec: fresnel}), and compare our method against a recent BRDF estimation approach (Sec.~\ref{sec: brdf}). In Sections~\ref{sec: data},\ref{sec: implementation}, we explain the data capture process and provide additional implementation details. Finally, we describe our supplementary video (Sec.~\ref{sec: video}), show additional novel-view synthesis results along with their intermediate rendering components (Sec.~\ref{sec: additional}).

\section{Effects of Loss Functions}\label{sec: loss}
In this section, we study how the choice of loss functions affects the quality of environment estimation and novel view synthesis. Specifically, we consider three loss functions between prediction and reference images as introduced in the main paper: (i) pixel-wise $L1$ loss, (ii) neural-network based perceptual loss, and (iii) adversarial loss. We run each of our algorithms (environment estimation and novel-view synthesis) for the three following cases: using (i) only, (i+ii) only, and all loss functions combined (i+ii+iii). For both algorithms we provide visual comparisons for each set of loss functions in Figures~\ref{fig: naive},\ref{fig: envmap}.
\subsection{Environment Estimation}
We run our joint optimization of SRMs and material weights to recover a visualization of the environment using the set of loss functions described above. As shown in Fig. \ref{fig: envmap}, the pixel-wise L1 loss was unable to effectively penalize the view prediction error because it is very sensitive to misalignments due to noisy geometry and camera pose. While the addition of perceptual loss produces better results, one can observe muted specular highlights in the very bright regions. The adversarial loss, in addition to the two other losses, effectively deals with the input errors while simultaneously correctly capturing the light sources.
\subsection{Novel-View Synthesis}
\begin{figure}
\hfill
\begin{subfigure}[t]{0.24\linewidth}
\adjincludegraphics[width=\linewidth,trim={{.15\width} 0 {.17\width} {.05\width}
},clip]{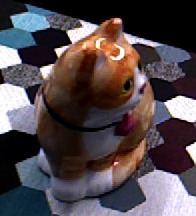}
\caption{GT}
\end{subfigure}
\hfill
\begin{subfigure}[t]{0.24\linewidth}
\adjincludegraphics[width=\linewidth,trim={{.15\width} 0 {.17\width} {.05\width}
},clip]{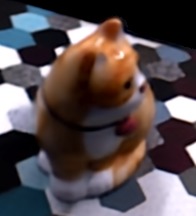}
\caption{L1 Loss}
\end{subfigure}
\hfill
\begin{subfigure}[t]{0.24\linewidth}
\adjincludegraphics[width=\linewidth,trim={{.15\width} 0 {.17\width} {.05\width}
},clip]{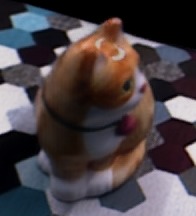}
\caption{L1+Percept}
\end{subfigure}
\hfill
\begin{subfigure}[t]{0.24\linewidth}
\adjincludegraphics[width=\linewidth,trim={{.15\width} 0 {.17\width} {.05\width}
},clip]{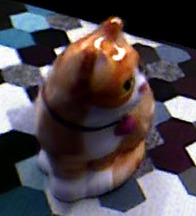}
\caption{All Losses}
\end{subfigure}
\hfill

\caption{Effects of loss functions on neural-rendering. The specular highlights on the forehead of the Labcat is expressed weaker than it actually is when using L1 or perceptual loss, likely due to geometric and calibration errors. The highlight is best expressed when the neural rendering pipeline of Sec. 6 is trained with the combination of L1, perceptual, and adversarial loss. 
} \label{fig: naive}
\end{figure}

\begin{figure*}
\hfill
\begin{subfigure}[t]{0.12\linewidth}
\includegraphics[width=\linewidth]{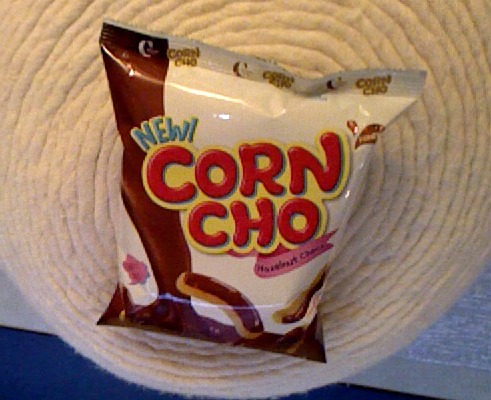}
\end{subfigure}
\hfill
\begin{subfigure}[t]{0.28\linewidth}
\includegraphics[width=\linewidth]{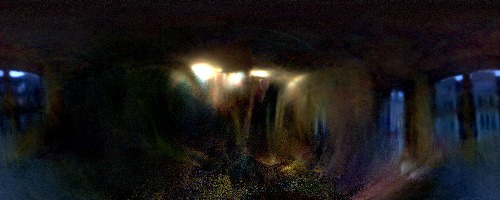}
\end{subfigure}
\hfill
\begin{subfigure}[t]{0.28\linewidth}
\includegraphics[width=\linewidth]{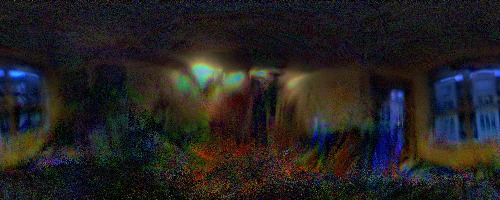}
\end{subfigure}
\hfill
\begin{subfigure}[t]{0.28\linewidth}
\includegraphics[width=\linewidth]{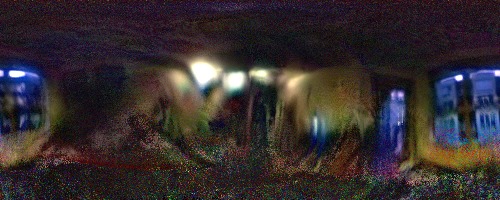}
\end{subfigure}
\hfill

\hfill
\begin{subfigure}[t]{0.12\linewidth}
\includegraphics[width=\linewidth]{images/labcat_recon_crop.jpg}
\caption{Scene}
\end{subfigure}
\hfill
\begin{subfigure}[t]{0.28\linewidth}
\includegraphics[width=\linewidth]{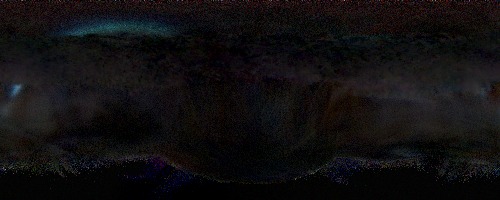}
\caption{L1 Loss}
\end{subfigure}
\hfill
\begin{subfigure}[t]{0.28\linewidth}
\includegraphics[width=\linewidth]{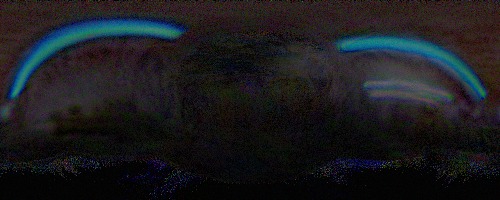}
\caption{L1+Perceptual Loss}
\end{subfigure}
\hfill
\begin{subfigure}[t]{0.28\linewidth}
\includegraphics[width=\linewidth]{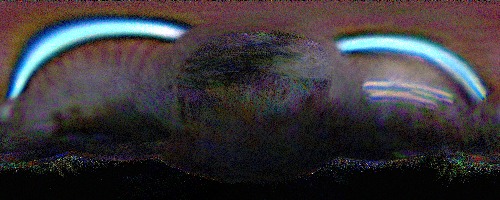}
\caption{L1+Perceptual+GAN Loss }
\end{subfigure}
\hfill

\caption{Environment estimation using different loss functions. From  input video sequences (a), we run our SRM estimation algorithm, varying the final loss function between the view predictions and input images. Because L1 loss (b) is very sensitive to misalignments caused by geometric and calibration errors, it averages out the observed specular highlights, resulting in missing detail for large portions of the environment. While the addition of perceptual loss (c) mitigates this problem, the resulting SRMs often lose the brightness or details of the specular highlights. The adoption of GAN loss produces improved results (d). } \label{fig: envmap}
\end{figure*}

We similarly train the novel-view neural rendering network in Sec. 6 using the aforementioned loss functions. Results in Fig.~\ref{fig: naive} shows that while L1 loss fails to capture specularity when significant image misalignments exist, the addition of perceptual loss somewhat addresses the issue. As expected, using adversarial loss, along with all other losses, allows the neural network to fully capture the intensity of specular highlights.

\section{Effects of Surface Roughness} \label{sec: roughness}
As descrbied in the main paper, our recovered specular reflectance map is environment lighting convolved with the surface's specular BRDF. Thus, the quality of the estimated SRM should depend on the roughness of the surface, e.g. a near Lambertian surface would not provide significant information about its surroundings. To test this claim, we run the SRM estimation algorithm on a synthetic object with varying levels of specular roughness. Specifically, we vary the roughness parameter of the GGX shading model \cite{walter2007microfacet} from 0.01 to 1.0, where smaller values correspond to more mirror-like surfaces. 
We render images of the synthetic object, and provide those rendered images, as well as the geometry (with added noise in both scale and vertex displacements, to simulate a real scanning scenario), to our algorithm. The results show that the accuracy of environment estimation decreases as the object surface gets more rough, as expected (Fig. \ref{fig:roughness}). Note that although increasing amounts of surface roughness does cause the amount of detail in our estimated environments to decrease, this is expected, as the recovered SRM still faithfully reproduces the convolved lighting (Fig. \ref{fig: rough_SRM}).

\section{Effects of Number of Material Bases} \label{sec: basis}
\begin{figure}
\hfill
\begin{subfigure}[t]{0.31\linewidth}
\includegraphics[width=\linewidth]{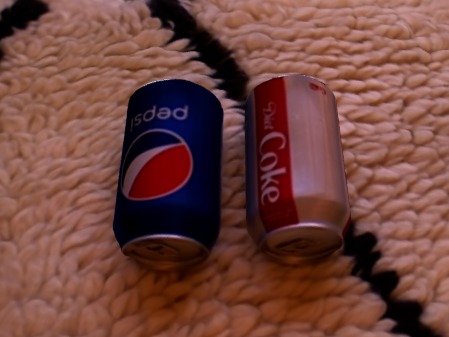}
\caption{Input Texture}
\end{subfigure}
\hfill
\begin{subfigure}[t]{0.31\linewidth}
\includegraphics[width=\linewidth]{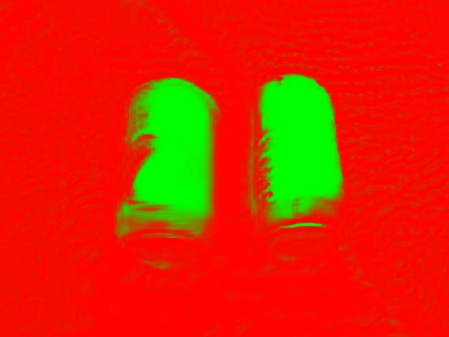}
\caption{Material Weight, $M=2$}
\end{subfigure}
\hfill
\begin{subfigure}[t]{0.31\linewidth}
\includegraphics[width=\linewidth]{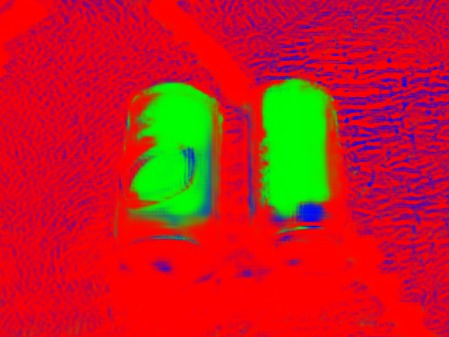}
\caption{Material Weight, $M=3$}
\end{subfigure}
\hfill

\hfill
\begin{subfigure}[t]{0.48\linewidth}
\includegraphics[width=\linewidth]{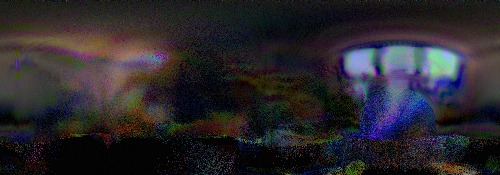}
\caption{Recovered SRM, $M=2$}
\end{subfigure}
\hfill
\begin{subfigure}[t]{0.48\linewidth}
\includegraphics[width=\linewidth]{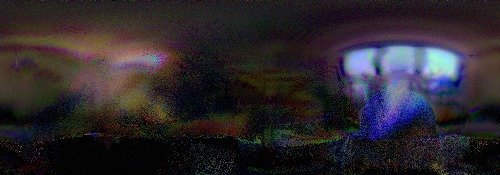}
\caption{Recovered SRM, $M=3$}
\end{subfigure}
\hfill

\caption{Sensitivity to the number of material bases $M$. We run our SRM estimation and material segmentation pipeline twice on a same scene but with different number of material bases $M$, showing that our system is robust to the choice of $M$. We show the predicted combination weights of the network trained with two (b) and three (c) material bases. For both cases (b,c), SRMs that correspond to the red and blue channel are mostly black, i.e. diffuse BRDF. Note that our algorithm consistently assigns the specular material (green channel) to the same regions of the image (cans), and that the recovered SRMs corresponding to the green channel (d,e) are almost identical.    }
\label{fig:material_sensitivity}
\end{figure}

The joint SRM and segmentation optimization of the main paper requires a user to set the number of material bases. In this section, we study how the algorithm is affected by the user specified number. Specifically, for a scene containing two cans, we run our algorithm twice, with number of material bases set to be two and three, respectively. The results of the experiment in Figure~\ref{fig:material_sensitivity} suggest that the number of material bases does not have a significant effect on the output of our system.

\section{Fresnel Effect Example} \label{sec: fresnel}
The Fresnel effect is a phenomenon where specular highlights tend to be stronger at near-glancing view angles, and is an important visual effect in the graphics community. We show in Fig.~\ref{fig: fresnel} that our neural rendering system correctly models the Fresnel effect. In the supplementary video, we show the Fresnel effect in motion, along with comparisons to the ground truth sequences.

\section{Comparison to BRDF Fitting} \label{sec: brdf}
Recovering a parametric analytical BRDF is a popular strategy to model view-dependent effects. We thus compare our neural network-based novel-view synthesis approach against a recent BRDF fitting method of \cite{park2018surface} that uses an IR laser and camera to optimize for the surface specular BRDF parameters. As shown in Fig.~\ref{fig: slfusion}, sharp specular BRDF fitting methods are prone to failure when there are calibration errors or misalignments in geometry. 

\begin{figure*}
\hfill
\begin{subfigure}[t]{0.20\linewidth}
\adjincludegraphics[width=\linewidth,trim={{.5\width} {0.15\width} {0.1\width} {.05\width}
},clip]{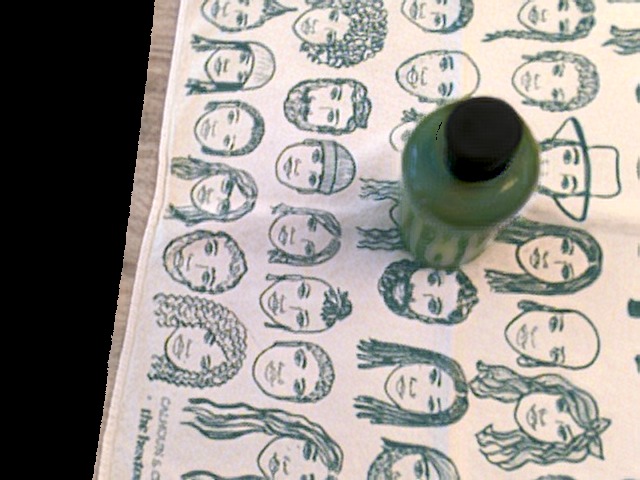}
\caption{View 1}
\end{subfigure}
\hfill 
\begin{subfigure}[t]{0.20\linewidth}
\adjincludegraphics[width=\linewidth,trim={{.45\width} {0.1\width} {0.15\width} {.1\width}
},clip]{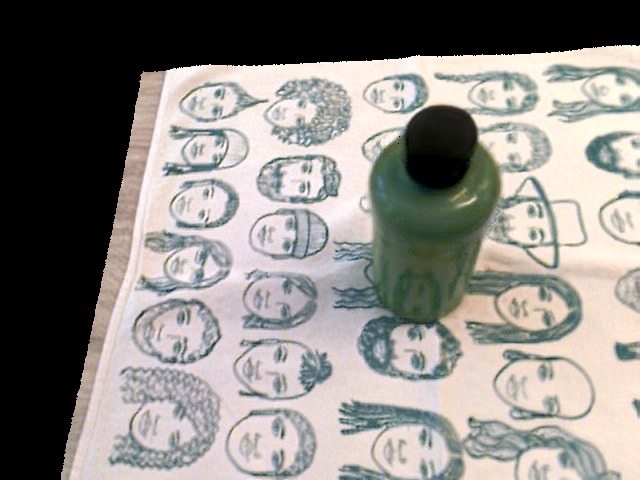}
\caption{View 2}
\end{subfigure}
\hfill
\begin{subfigure}[t]{0.20\linewidth}
\adjincludegraphics[width=\linewidth,trim={{.5\width} {0.15\width} {0.1\width} {.05\width}
},clip]{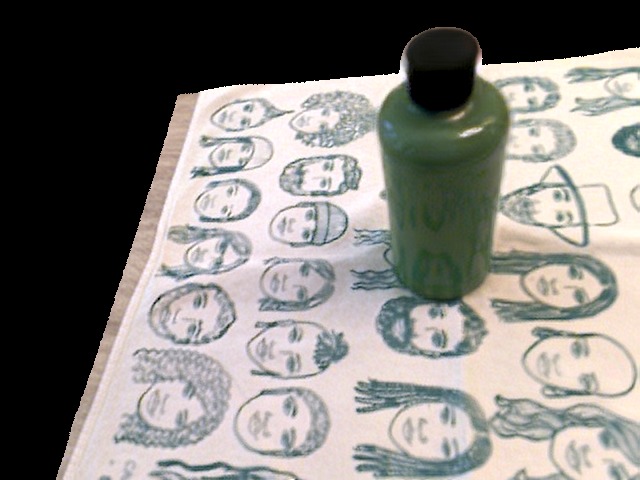}
\caption{View 3}
\end{subfigure}
\hfill
\begin{subfigure}[t]{0.12\linewidth}
\adjincludegraphics[width=\linewidth,trim={{.5\width} {0.15\width} {0.1\width} {.05\width}
},clip]{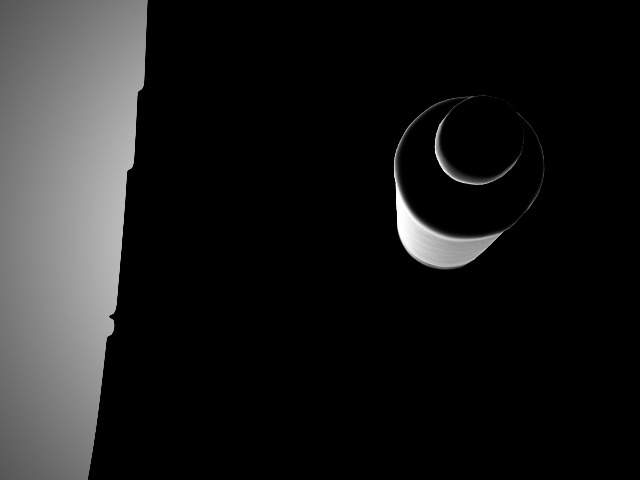}
\caption{View 1}
\end{subfigure}
\hfill 
\begin{subfigure}[t]{0.12\linewidth}
\adjincludegraphics[width=\linewidth,trim={{.45\width} {0.1\width} {0.15\width} {.1\width}
},clip]{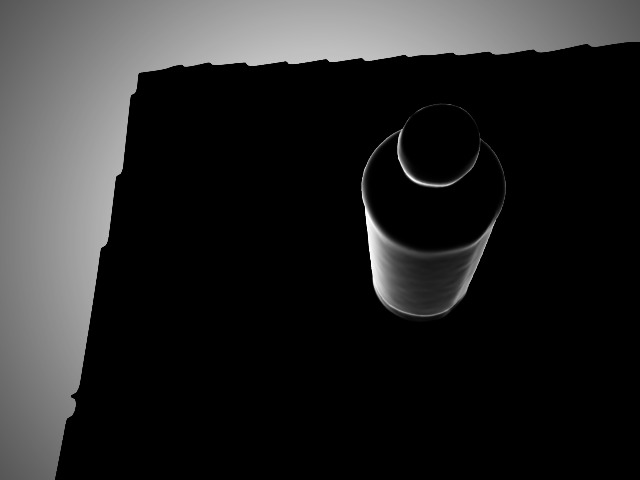}
\caption{View 2}
\end{subfigure}
\hfill
\begin{subfigure}[t]{0.12\linewidth}
\adjincludegraphics[width=\linewidth,trim={{.5\width} {0.15\width} {0.1\width} {.05\width}
},clip]{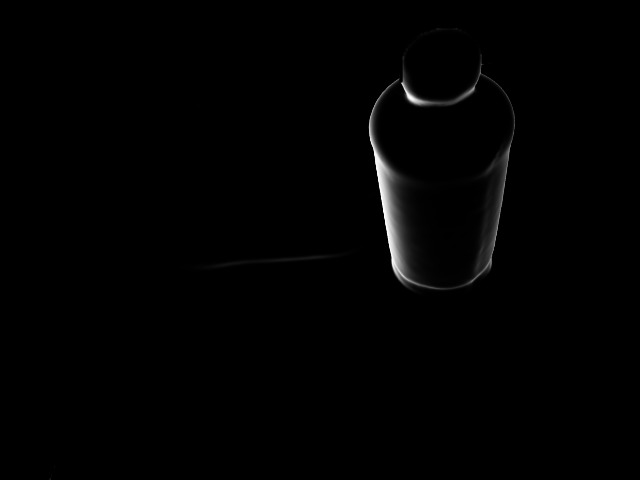}
\caption{View 3}
\end{subfigure}
\hfill

\caption{Demonstration of the Fresnel effect. The intensity of specular highlights tends to be amplified at slant viewing angles. We show three different views (a,b,c) for a glossy bottle, each of them generated by our neural rendering pipeline and presenting different viewing angles with respect to the bottle. Notice that the neural rendering correctly amplifies the specular highlights as the viewing angle gets closer to perpendicular with the surface normal. Images (d,e,f) show the computed Fresnel coefficient (FCI) (see Sec. 6.1) for the corresponding views. These images are given as input to the neural-renderer that subsequently use them to simulate the Fresnel effect. Best viewed digitally.
} \label{fig: fresnel}
\end{figure*} 

\begin{figure} 

\centering
\begin{subfigure}[t]{0.7\linewidth}
\adjincludegraphics[width=\linewidth,trim={{.0\width} {0.14\width} {0.\width} {.0\width}
},clip]{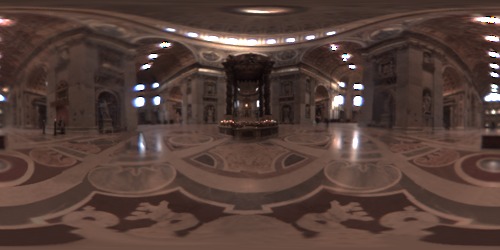}
\caption{Ground Truth Environment}
\end{subfigure}

\hfill
\begin{subfigure}[t]{0.28\linewidth}
\adjincludegraphics[width=\linewidth,trim={{.24\width} {0.14\width} {0.24\width} {.14\width}
},clip]{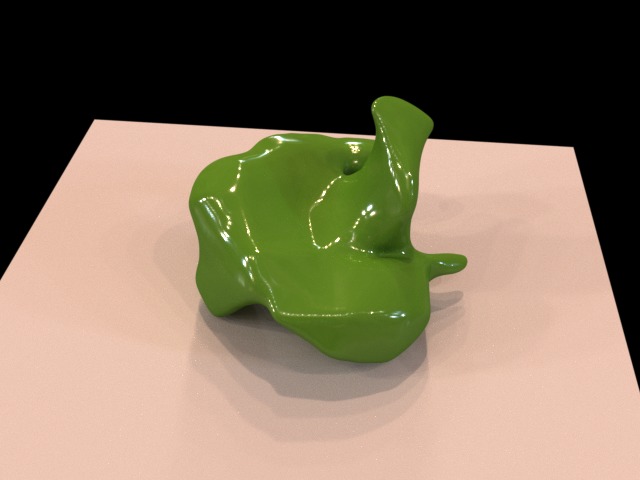}
\caption{Input Frame}
\end{subfigure}  
\hfill
\begin{subfigure}[t]{0.70\linewidth}
\adjincludegraphics[width=\linewidth,trim={{.0\width} {0.14\width} {0.\width} {.0\width}
},clip]{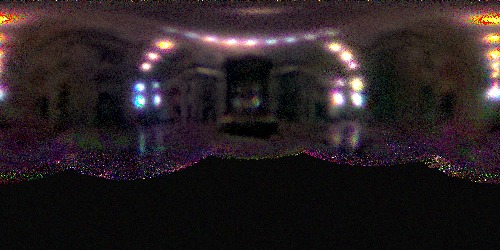}
\caption{Recovered SRM (GGX roughness 0.01)}
\end{subfigure}
\hfill

\hfill
\begin{subfigure}[t]{0.28\linewidth}
\adjincludegraphics[width=\linewidth,trim={{.24\width} {0.14\width} {0.24\width} {.14\width}
},clip]{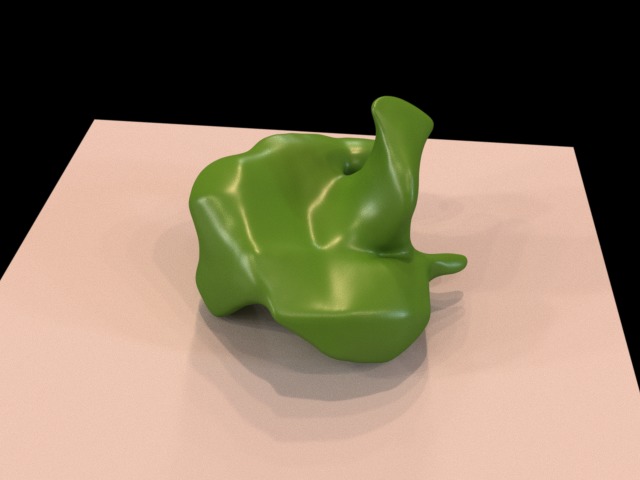}
\caption{Input Frame}
\end{subfigure}  
\hfill
\begin{subfigure}[t]{0.70\linewidth}
\adjincludegraphics[width=\linewidth,trim={{.0\width} {0.14\width} {0.\width} {.0\width}
},clip]{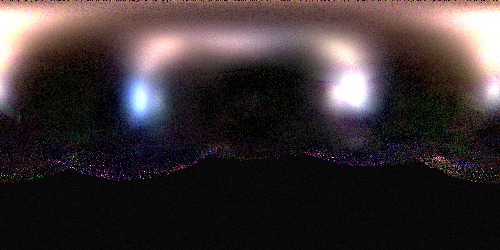}
\caption{Recovered SRM (GGX roughness 0.1)}
\end{subfigure}
\hfill

\hfill
\begin{subfigure}[t]{0.28\linewidth}
\adjincludegraphics[width=\linewidth,trim={{.24\width} {0.14\width} {0.24\width} {.14\width}
},clip]{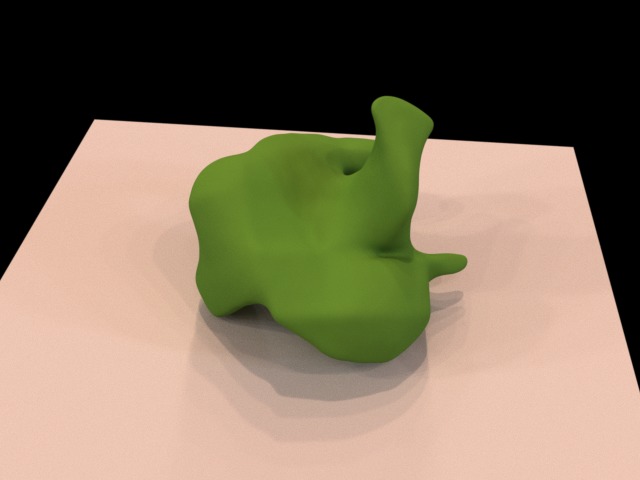}
\caption{Input Frame}
\end{subfigure} 
\hfill
\begin{subfigure}[t]{0.70\linewidth}
\adjincludegraphics[width=\linewidth,trim={{.0\width} {0.14\width} {0.\width} {.0\width}
},clip]{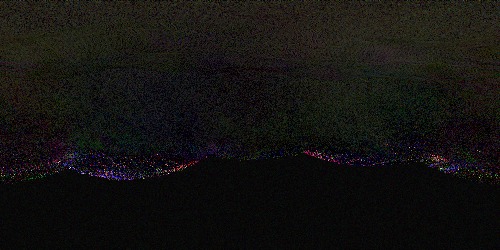}
\caption{Recovered SRM (GGX roughness 0.7)}
\end{subfigure}
\hfill

\caption{Recovering SRM for different surface roughness. We test the quality of estimated SRMs (c,e,g) for various surface materials (shown in (b,d,f)). The results closely match our expectation that environment estimation through specularity is challenging for glossy (d) and diffuse (f) surfaces, compared to the mirror-like surfaces (c). Note that the input to our system are rendering images and noisy geometry, from which our system reliably estimates the environment. }\label{fig: rough_SRM}
\end{figure}

\begin{figure}
    \centering
    \includegraphics[width=\linewidth]{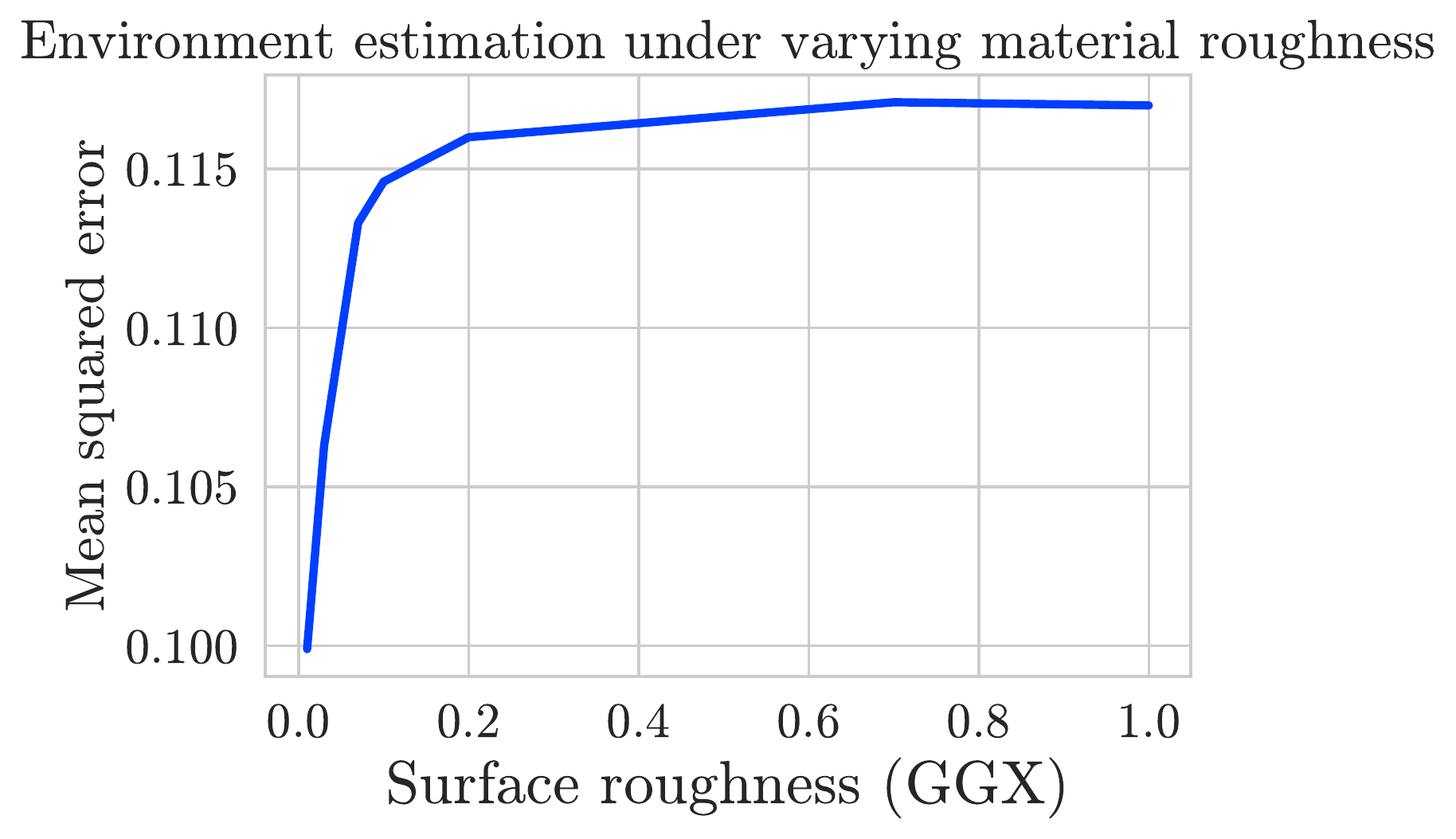}
    \caption{Accuracy of environment estimation under dif
ferent amounts of surface roughness. We see that increas
ing the material roughness does indeed decrease the over
all quality of the reconstructed environment image measured in pixel-wise L2 distance. Note
that the roughness parameter is from the GGX \cite{walter2007microfacet} shading
model which we use to render the synthetic models.}
    \label{fig:roughness}
\end{figure}

\section{Data Capture Details} \label{sec: data}
As described in Sec. 7 of the main paper, we capture ten videos of objects with varying materials, lighting and compositions. We used a Primesense Carmine RGBD structured light camera. We perform intrinsic and radiometric calibrations, and correct the images for vignetting. During capture, the color and depth streams were hardware-synchronized, and registered to the color camera frame-of-reference. The resolution of both streams are VGA (640x480) and the frame rate was set to 30fps. Camera exposure was manually set and fixed within a scene. 

We obtained camera extrinsics by running ORB-SLAM \cite{mur2017orb} (ICP \cite{newcombe2011kinectfusion} was alternatively used for feature-poor scenes). Using the estimated pose, we ran volumetric fusion \cite{newcombe2011kinectfusion} to obtain the geometry reconstruction. Once geometry and rough camera poses are estimated, we ran frame-to-model dense photometric alignment following \cite{park2018surface} for more accurate camera positions, which are subsequently used to fuse in the diffuse texture to the geometry. Following \cite{park2018surface}, we use iteratively reweighted least squares to compute a robust minimum of intensity for each surface point across viewpoints, 
which provides a good approximation to the diffuse texture.

\section{Implementation Details} \label{sec: implementation}
Our pipeline is built using PyTorch \cite{paszke2017automatic}. For all of our experiments we used ADAM optimizer with learning rate 2e-4 for the neural networks and 1e-3 for the SRM pixels. For the SRM optimization described in Sec. 5 of the main text the training was run for 40 epochs (i.e. each training frame is processed 40 times), while the neural renderer training was run for 75 epochs.

We find that data augmentation plays a significant role to the view generalization of our algorithm. For training in Sec. 5, we used random rotation (up to 180$^\circ$), translation (up to 100 pixels), and horizontal and vertical flips. For neural renderer training in Sec. 6, we additionally scale the input images by a random factor between 0.8 and 1.25.

We use Blender \cite{blender} for computing the reflection direction image $R_P$ and the first bounce interreflection (FBI) image described in the main text.

\subsection{Network Architectures}
Let \texttt{C(k,ch\_in,ch\_out,s)} be a convolution layer with kernel size \texttt{k}, input channel size \texttt{ch\_in}, output channel size \texttt{ch\_out}, and stride \texttt{s}. When the stride \texttt{s} is smaller than 1, we first conduct nearest-pixel upsampling on the input feature and then process it with a regular convolution layer. We denote \texttt{CNR} and \texttt{CR} to be the Convolution-InstanceNorm-ReLU layer and Convolution-ReLU layer, respectively. 
A residual block \texttt{R(ch)} of channel size \texttt{ch} contains convolutional layers of \texttt{CNR(3,ch,ch,1)-CN(3,ch,ch,1)}, where the final output is the sum of the outputs of the first and the second layer.

\paragraph{Encoder-Decoder Network Architecture}
The architecture of the texture refinement network and the neural rendering network in Sec.5 and Sec.6 closely follow the architecture of an encoder-decoder network of Johnson \etal \cite{johnson2016perceptual}: \texttt{CNR(9,ch\_in,32,1)-CNR(3,32,64,2)-CNR(3,64,\\128,2)-R(128)-R(128)-R(128)-R(128)-R(128)\\-CNR(3,128,64,1/2)-CNR(3,64,32,1/2)\\-C(3,32,3,1)}, where \texttt{c\_in} represents a variable input channel size, which is 3 and 13 for the texture refinement network and neural rendering generator, respectively.

\paragraph{Material Weight Network}
The architecture of the material weight estimation network in Sec. 5 is as follows: \texttt{CNR(5,3,64,2)-CNR(3,64,64,2)-R(64)-R(64)-\\CNR(3,64,32,1/2)-C(3,32,3,1/2)}.

\paragraph{Discriminator Architecture}
The discriminator network used for the adversarial loss in Eq.7 and Eq.8 of the main paper both use the same architecture as follows: \texttt{CR(4,3,64,2)-CNR(4,64,128,2)-CNR(4,128,\\256,2)-CNR(4,256,512,2)-C(1,512,1,1)}. For this network, we use a LeakyReLU activation (slope 0.2) instead of the regular ReLU, so CNR used here is a Convolution-InstanceNorn-LeakyReLU layer. Note that the spatial dimension of the discriminator output is larger than 1x1 for our image dimensions (640x480), i.e., the discriminator scores realism of patches rather than the whole image (as in PatchGAN \cite{isola2017image}).
\let\thefootnote\relax\footnotetext{$^\dagger$\text{Video URL: }\url{https://youtu.be/9t_Rx6n1HGA}}

\section{Supplementary Video} \label{sec: video}
We strongly encourage readers to watch the supplementary video$^\dagger$, as many of our results we present are best seen as videos. Our supplementary video contains visualizations of input videos, environment estimations, our neural novel-view synthesis (NVS) renderings, and side-by-side comparisons against the state-of-the-art NVS methods. We note that the ground truth videos of the NVS section are cropped such that regions with missing geometry are displayed as black. The purpose of the crop is to provide equal visual comparisons between the ground truth and the rendering, so that viewers are able to focus on the realism of reconstructed scene instead of the background. Since the reconstructed geometry is not always perfectly aligned with the input videos, some boundaries of the ground truth stream may contain noticeable artifacts, such as edge-fattening. An example of this can be seen in the `acryl' sequence, near the top of the object. 

\section{Additional Results} \label{sec: additional}
\begin{table}[]
\centering
\begin{tabular}{rcccc}
\multicolumn{1}{l}{}              & Cans-L1                       & Labcat-L1        & Cans-perc  &Labcat-perc           \\ \cline{2-5} 
\multicolumn{1}{r|}{\cite{hedman2018deep}} & \multicolumn{1}{c|}{9.82e-3} & \multicolumn{1}{c|}{6.87e-3} & \multicolumn{1}{c|}{0.186} & \multicolumn{1}{c|}{0.137} \\ \cline{2-5} 
\multicolumn{1}{r|}{\cite{thies2019deferred}}  & \multicolumn{1}{c|}{9.88e-3} & \multicolumn{1}{c|}{8.04e-3} & \multicolumn{1}{c|}{0.163} & \multicolumn{1}{c|}{0.178}\\ \cline{2-5} 
\multicolumn{1}{r|}{Ours}         & \multicolumn{1}{c|}{\bf 4.51e-3} & \multicolumn{1}{c|}{\bf 5.71e-3} & \multicolumn{1}{c|}{\bf 0.103} & \multicolumn{1}{c|}{\bf 0.098}\\ \cline{2-5} 
\end{tabular}
\hfill
\caption{Average pixel-wise L1 error and perceptual error values (lower is better) across the different view synthesis methods on the two datasets (Cans, Labcat). The L1 metric is computed as mean L1 distance across pixels and channels between novel-view prediction and ground-truth images. The perceptual error numbers correspond to the mean values of the measurements shown in Figure~7 of the main paper. As described in the main paper, we mask out the background (e.g. carpet) and focus only on the specular object surfaces.} \label{tbl:mean_perceptual}
\end{table}

Table \ref{tbl:mean_perceptual} shows numerical comparisons on novel-view synthesis against state-of-the-art methods \cite{hedman2018deep,thies2019deferred} for the two scenes presented in the main text (Fig. 7). We adopt two commonly used  metrics, i.e. pixel-wise L1 and deep perceptual loss \cite{johnson2016perceptual}, to measure the distance between a predicted novel-view image and its corresponding ground-truth test image held-out during training.  As described in the main text we focus on the systems' ability to extrapolate specular highlight, thus we only measure the errors on the object surfaces, i.e. we remove diffuse backgrounds.

\begin{figure} 
\begin{subfigure}[t]{0.30\linewidth}
\includegraphics[width=\linewidth]{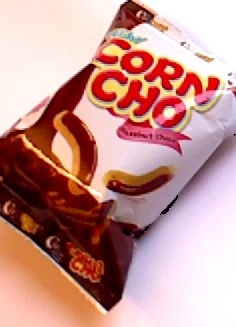}
\caption{Reference}
\end{subfigure}
\begin{subfigure}[t]{0.30\linewidth}
\includegraphics[width=\linewidth]{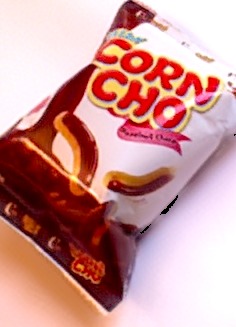}
\caption{Our Reconstruction}
\end{subfigure}
\begin{subfigure}[t]{0.30\linewidth}
\includegraphics[width=\linewidth]{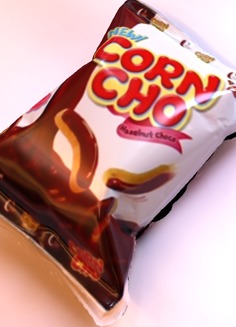}
\caption{Reconstruction by \cite{park2018surface}}
\end{subfigure}
	\caption{Comparison with Surface Light Field Fusion \cite{park2018surface}. Note that the sharp specular highlight on the bottom-left  of the Corncho bag is poorly reconstructed in the rendering of \cite{park2018surface} (c). As shown in Sec.~\ref{sec: loss} and Fig.~\ref{fig: main}, these high frequency appearance details are only captured when using neural rendering and robust loss functions (b).
	}
	\label{fig: slfusion}
\end{figure}

\begin{figure} 
\hfill
\begin{subfigure}[t]{0.45\linewidth}
\adjincludegraphics[width=\linewidth,trim={{.15\width} {0.15\width} {0.3\width} {.15\width}
},clip]{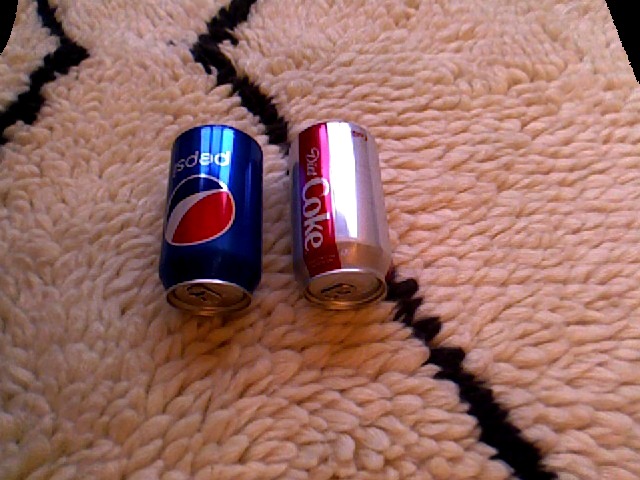}
\caption{Ground Truth}
\end{subfigure} 
\hfill
\begin{subfigure}[t]{0.45\linewidth}
\adjincludegraphics[width=\linewidth,trim={{.15\width} {0.15\width} {0.3\width} {.15\width}
},clip]{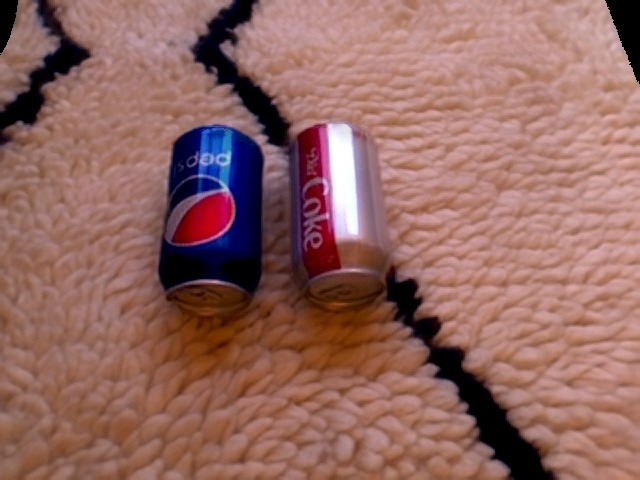}
\caption{Rendering with SRM}
\end{subfigure}
\hfill
	\caption{Motivation for neural rendering. While the SRM and segmentation obtained from the optimization of Sec. 5 of the main text provides high quality environment reconstruction, the simple addition of the diffuse and specular component does not yield photorealistic rendering (b) compared to the ground truth (a). This motivates the neural rendering network that takes input as the intermediate rendering components and generate photorealistic images (e.g. shown in Fig.~\ref{fig: main}).
	}
	\label{fig: motivation}
\end{figure}

Fig. \ref{fig: motivation} shows that the na\"ive addition of diffuse and specular components obtained from the optimization in Sec. 5 does not results in photorealistic novel view synthesis, thus motivating a separate neural rendering step that takes as input the intermediate physically-based rendering components.

\begin{figure*} 
\begin{subfigure}[t]{\linewidth}

\includegraphics[width=0.246\linewidth]{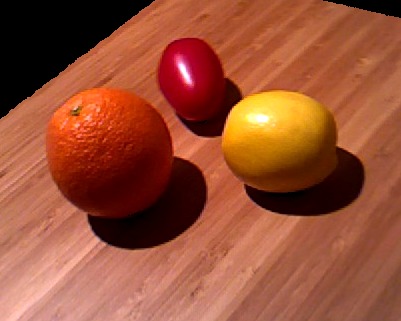}
\hfill
\includegraphics[width=0.246\linewidth]{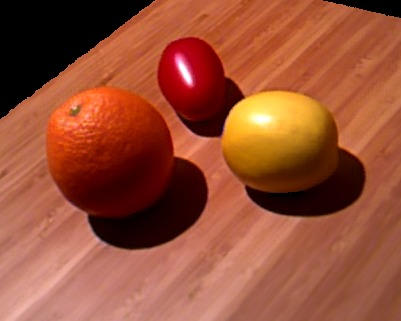}
\hfill
\includegraphics[width=0.246\linewidth]{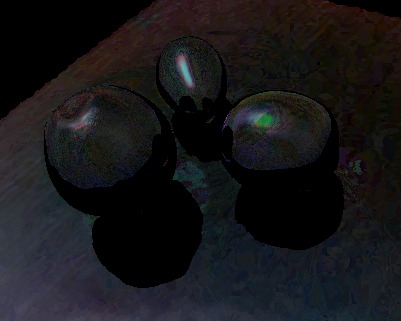}
\hfill
\includegraphics[width=0.246\linewidth]{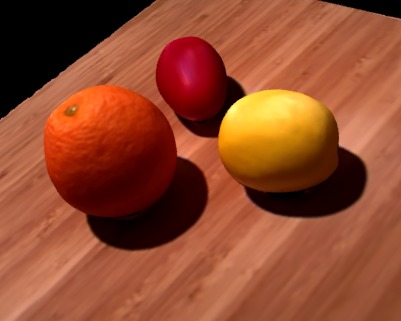}

\vspace{0.1cm}

\includegraphics[width=0.246\linewidth]{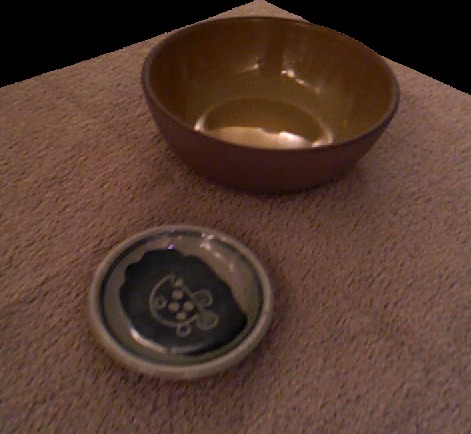}
\hfill
\includegraphics[width=0.246\linewidth]{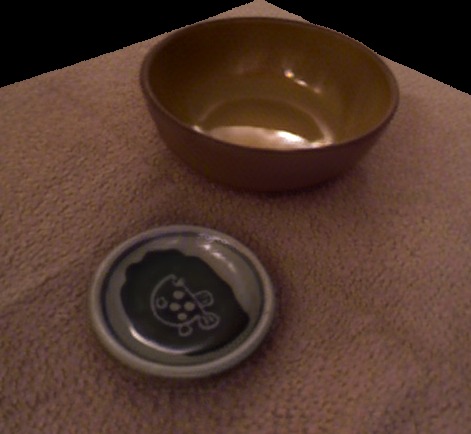}
\hfill
\includegraphics[width=0.246\linewidth]{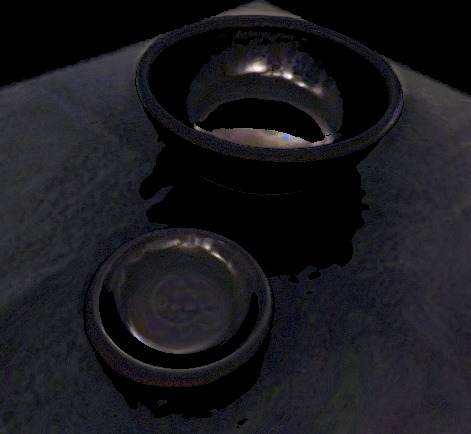}
\hfill
\includegraphics[width=0.246\linewidth]{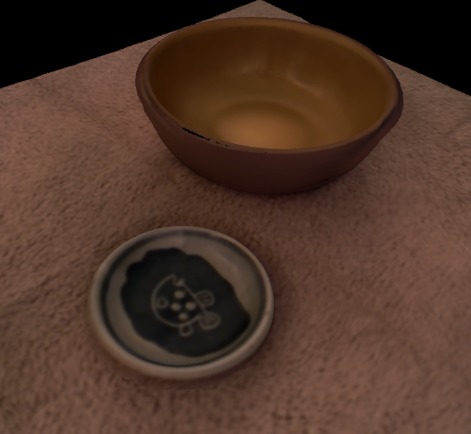}

\vspace{0.1cm}

\includegraphics[width=0.246\linewidth]{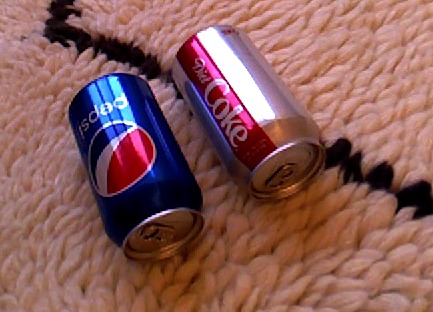}
\hfill
\includegraphics[width=0.246\linewidth]{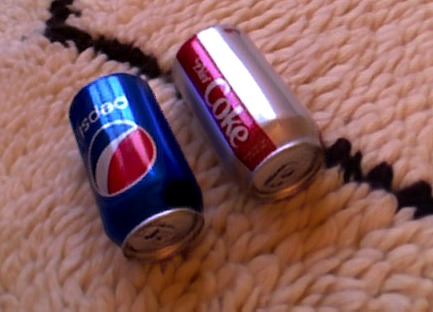}
\hfill
\includegraphics[width=0.246\linewidth]{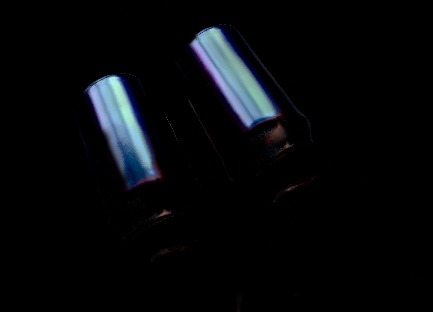}
\hfill
\includegraphics[width=0.246\linewidth]{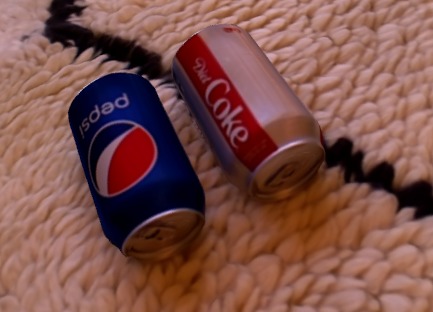}

\vspace{0.1cm}

\includegraphics[width=0.246\linewidth]{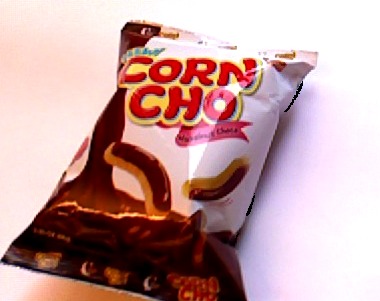}
\hfill
\includegraphics[width=0.246\linewidth]{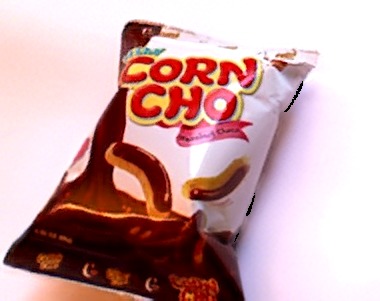}
\hfill
\includegraphics[width=0.246\linewidth]{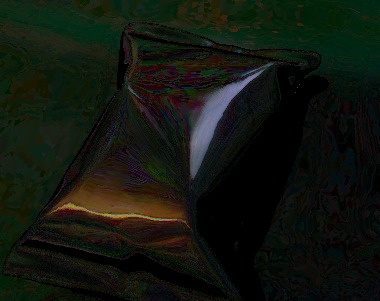}
\hfill
\includegraphics[width=0.246\linewidth]{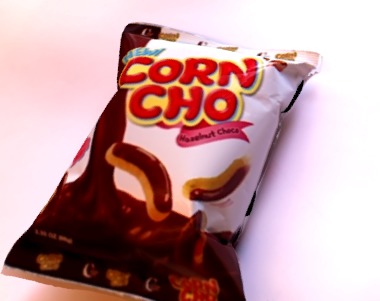}

\vspace{0.1cm}

\begin{subfigure}[t]{0.246\linewidth}
\includegraphics[width=\linewidth]{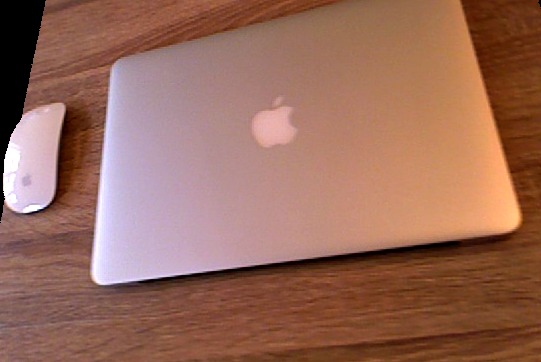}
\caption{Ground Truth $G_P$}
\end{subfigure}
\hfill
\begin{subfigure}[t]{0.246\linewidth}
\includegraphics[width=\linewidth]{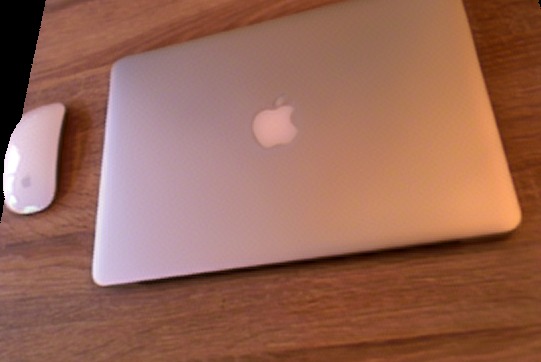}
\caption{Our Rendering $g(C_P)$}
\end{subfigure}
\hfill
\begin{subfigure}[t]{0.246\linewidth}
\includegraphics[width=\linewidth]{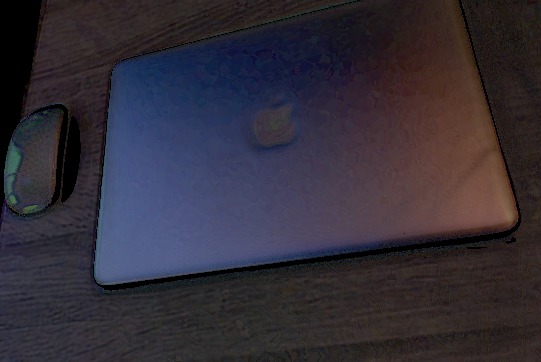}
\caption{Specular Component $S_P$}
\end{subfigure}
\hfill
\begin{subfigure}[t]{0.246\linewidth}
\includegraphics[width=\linewidth]{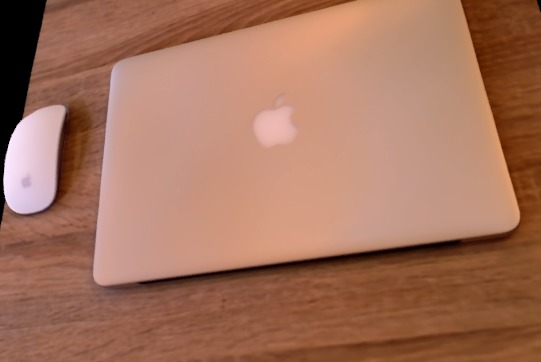}
\caption{Diffuse Component $D_P$}
\end{subfigure}

\end{subfigure}

\caption{Novel view renderings and intermediate rendering components for various scenes. From left to right: (a) reference photograph, (b) our rendering, (c) specular reflectance map image, and (d) diffuse texture image. Note that some of the ground truth reference images have black ``background'' pixels inserted near the top and left borders where reconstructed geometry is missing, to provide equal visual comparisons to rendered images. }\label{fig: main}
\end{figure*}

Fig.~\ref{fig: main} shows novel-view neural rendering results, together with the estimated components (diffuse and specular images $D_P$, $S_P$) provided as input to the renderer. Our approach can synthesize photorealistic novel views of a scene with wide range of materials, object compositions, and lighting condition. Note that the featured scenes contain challenging properties such as bumpy surfaces (Fruits), rough reflecting surfaces (Macbook), and concave surfaces (Bowls). Overall, we demonstrate the robustness of our approach for various materials including fabric, metals, plastic, ceramic, fruit, wood, glass, etc. 

On a separate note, reconstructing SRMs of planar surfaces could require more views to fully cover the environment hemisphere, because the surface normal variation of each view is very limited for a planar surface. We refer readers to Janick \etal \cite{jachnik2012real} that studies capturing planar surface light field, which reports that it takes about a minute using their real-time, guided capture system.

\end{document}